\documentclass[lettersize,journal]{IEEEtran}
\usepackage{amsmath,amsfonts}
\usepackage{algorithm}
\usepackage{array}
\usepackage[caption=false,font=normalsize,labelfont=sf,textfont=sf]{subfig}
\usepackage{textcomp}
\usepackage{stfloats}
\usepackage{url}
\usepackage{verbatim}
\usepackage{graphicx}
\usepackage{cite}
\usepackage[utf8]{inputenc} 
\usepackage[T1]{fontenc}    
\usepackage{url}            
\usepackage{booktabs}       
\usepackage{amsfonts}       
\usepackage{nicefrac}       
\usepackage{microtype}      
\usepackage{xcolor}         

\usepackage{amsmath}
\usepackage{amssymb}
\usepackage{mathtools}
\usepackage{amsthm}
\usepackage{setspace}
\theoremstyle{plain}

\theoremstyle{definition}

\theoremstyle{remark}

\usepackage{paperpkg}
\usepackage{algorithm}
\usepackage{algpseudocode}

\usepackage{wrapfig}

\let\ieeeappendix\appendix
\usepackage[capitalize,noabbrev]{cleveref}
\let\appendix\ieeeappendix

\usepackage{microtype}
\usepackage{graphicx}
\usepackage{booktabs} 
\usepackage{float}

\usepackage{amsmath}
\usepackage{amssymb}
\usepackage{mathtools}
\usepackage{amsthm}
\usepackage{setspace}

\usepackage[textsize=tiny]{todonotes}

\usepackage{paperpkg}
\usepackage{soul}

\newcommand{\alg}{{{MLP Fusion}}}

\newcommand{\yc}[1]{\textcolor{teal}{[YF: #1]}}

\newcommand{\ma}[1]{\textcolor{magenta}{[MT: #1]}}

\newcommand{\mytitle}{MLP Fusion: Towards Efficient Fine-tuning of Dense and Mixture-of-Experts Language Models}

\title{\mytitle}

\author{Mengting Ai$^{*}$, Tianxin Wei$^{*}$, Yifan Chen$^{*}$,~\IEEEmembership{Member,~IEEE}, Zeming Guo, Jingrui He,~\IEEEmembership{Senior Member,~IEEE}

\thanks{Mengting Ai, Tianxin Wei, and Jingrui He are with the School of Information Sciences, University of Illinois Urbana-Champaign, Champaign, IL 61820, USA. E-mail: mai10@illinois.edu; twei10@illinois.edu; jingrui@illinois.edu.}
\thanks{Yifan Chen is with the Departments of Mathematics and Computer Science, Hong Kong Baptist University, Kowloon, Hong Kong. E-mail: yifanc@hkbu.edu.hk.}
\thanks{Zeming Guo is with the Jacobs Technion-Cornell Institute, Cornell Tech, New York, NY 10044, USA. E-mail: zg296@cornell.edu.}
\thanks{*Mengting Ai, Tianxin Wei, and Yifan Chen contributed equally to this work.}
\thanks{Corresponding authors: Yifan Chen and Jingrui He.}}

\IEEEpubid{0000--0000/00\$00.00~\copyright~2021 IEEE}

\begin{document}

\maketitle

\begin{abstract}
Fine-tuning a pre-trained language model (PLM) emerges as the predominant strategy in many natural language processing applications. However, this process is known to be expensive, especially on edge devices with low computing power. While general approaches (e.g.\ quantization and distillation) have been widely studied to reduce the compute/memory of PLM fine-tuning, one-shot compression techniques specifically designed for fine-tuning remain largely unexplored. In this paper, we investigate the neural tangent kernel (NTK)--which reveals the gradient descent dynamics of neural networks--of the multilayer perceptrons (MLP) modules in a PLM and propose to coin a lightweight PLM through NTK-approximating \textit{MLP fusion}. By incorporating NTK into the compression process, \alg~not only preserves the original model's output but also maintains its training dynamics. To achieve this, we reconsider the MLP as a bundle of sub-MLPs and cluster them into a given number of centroids, which can then be restored as a compressed MLP and surprisingly well approximate the NTK of the original PLM. Our approach is applicable to both standard MLP modules and Mixture-of-Experts (MoE) modules in PLMs, demonstrating its scalability and versatility. Additionally, we provide theoretical derivations to demonstrate how the proposed compression preserves the NTK. Extensive experiments of PLM fine-tuning on both natural language understanding and generation tasks are provided to verify the effectiveness of MLP fusion. 
Our code is available at \url{https://github.com/weitianxin/MLP_Fusion}.

\end{abstract}


\section{Introduction}

\IEEEPARstart{S}{upervised} fine-tuning (SFT) of
pre-trained language models (PLMs) has been the most common method to tackle downstream natural language processing (NLP) tasks \cite{howard2018universal, kale-rastogi-2020-text}.
However, despite the high performance of SFT on downstream tasks
\cite{peters2018deep}, users face significant computational costs in terms of both time and space due to the large size of PLMs. The sizes of popular PLMs have recently grown from hundreds of millions~\cite{brown2020language} to trillions~\cite{fedus2022switch} of parameters, driven by scaling laws~\cite{kaplan2020scalinglawsneurallanguage}. Even the smallest BERT model~\cite{devlin2018bert} has over 110M parameters, not to mention the newer Llama-series models~\cite{touvron2023llama}, which range from 7B to 405B parameters.
Mixture-of-Experts (MoE) \cite{shazeer2017outrageously}, as another product of scaling laws, extends beyond the traditional feedforward neural network (FFN) layer by replacing a single multilayer perceptron (MLP) with multiple MLPs, referred to as ``experts''. 
Sparse MoE designs improve performance while keeping inference computational costs (FLOPs) comparable to those of the original dense model, as only a few selected experts are activated during inference. However, during fine-tuning, the computational costs significantly increase because each expert requires tuning. The expert size for Mixtral \cite{jiang2024mixtral} reaches 176.2M, and the presence of 8 or even more experts in each layer exacerbates the demands.

Various efforts have been made in various fields to compress and harness the large-scale PLMs.
A popular technique in \textit{model compression} is knowledge distillation~\cite[KD]{hinton2015distilling,gou2021knowledge,10.1145/3534678.3539315,10.1145/3340531.3412005}, 
which aims to transfer knowledge from pre-trained large language models (LLMs) to smaller models. However, this approach requires extensive retraining, involving both the original LLM and the compact model.
There were some previous attempts to establish one-shot model compression methods.
Single-shot \textit{pruning} methods \cite[sparsification]{han2015learning, lee2018snip, wang2020picking, tanaka2020pruning} identify sub-networks at initialization concerning certain criteria of weight magnitude or gradient flow.
However, most works on (entry-wise) pruning focus on reducing the conceptual training or inference FLOPs, while sparse matrix multiplication is not well supported on modern hardware (e.g.\ GPUs) and even slows down the training in wall-clock time~\cite{dao2022monarch}.

\begin{figure}[t]
    \centering
    \includegraphics[width=1\columnwidth]{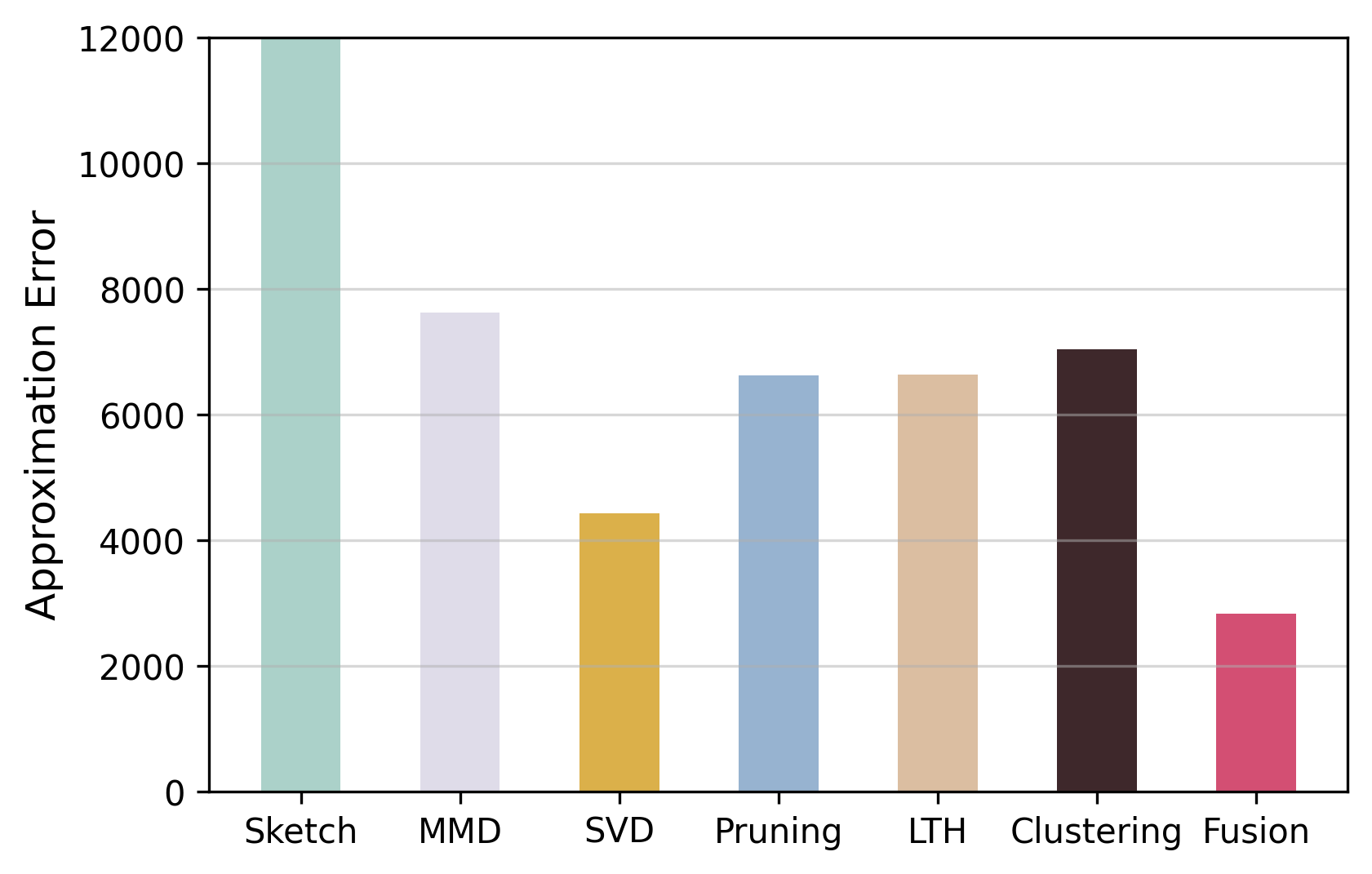}
    \caption{NTK matrix approximation error of compression methods on the validation set of SST2.}
    \label{fig:ntk-error}
\vspace{-0.15in}
\end{figure}

\IEEEpubidadjcol

Meanwhile, truncated singular value decomposition (SVD) on weight matrices \cite{denton2014exploiting,wang2024svdllmtruncationawaresingularvalue} has been applied to accelerate large CNNs by leveraging the linear structure within the network to eliminate redundancy. However, truncated SVD has limited representational power and might lead to suboptimal performance, as it significantly reduces the dimensionality of the linear transformations in the network. LoRA~\cite{hu2021loralowrankadaptationlarge}, although mitigating this issue by incorporating information from the original weight matrices, does not reduce the inference cost of the fine-tuned model. Specifically for MoE models, various studies have introduced the concept of expert merging \cite{he2023merging,li2023merge,xue2022student,pmlr-v162-liu22k,stoica2024zipitmergingmodelsdifferent} and expert pruning \cite{lu2024expertsequalefficientexpert}, as a method to reduce the number of experts within each layer of the MoE model. 

On a separate note, although efficient attention mechanisms~\cite{DBLP:conf/iclr/KitaevKL20,DBLP:journals/corr/abs-2009-14794,chen2021skyformer} have become the mainstream methodology to accelerate the pre-training of language models, we instead focus on the
FFN sub-layers of a pre-trained transformer.
Given the ever-increasing hidden input size of PLMs, which now exceeds ten thousand \cite{touvron2023llama,jiang2024mixtral}, the significance of FFN sub-layers in terms of computation cost has grown substantially. We observe for regular NLP tasks where the token sequence length is no longer than $512$, the computational cost of the MLP modules is heavier than the attention module even though the attention module has a quadratic complexity (detailed computation and comparison of the computational cost within the two modules are provided in the supplementary material Appendix B-A). This disparity is even more pronounced in MoE models.

The current limitations of general model compression methods and the realistic need to reduce the compute in MLP modules motivate us to develop an MLP compression technique for efficient language model SFT.
To attain competitive SFT performance, we propose a novel perspective on PLM compression, that the compressed model is supposed to approximate not only the model output, but also the training dynamics of the original SFT. 
We turn to neural tangent kernel~\cite[NTK]{jacot2018neural, arora2019exact} as a proxy of the SFT dynamics and manage to enable the compressed model to approximate the original NTK;
specifically, we dissect the MLP or MoE structure in a PLM, connect it with model fusion~\cite[which layer-wisely fuse multiple MLPs into one]{singh2020model}, 
and propose a novel compression method, \textit{MLP fusion}, specific to PLM fine-tuning.
As shown in \Cref{fig:ntk-error}, our method ``fusion'' provably attains the smallest NTK matrix approximation error on a real-world dataset SST2~\cite{socher2013recursive}.

In summary, the contributions of this work are four-fold: \begin{itemize}[leftmargin=*]
\item We introduce the concept of NTK approximation for PLM compression, to ensure that the compressed model can preserve the training dynamics of the original model.
\item We dissect the MLP modules in PLMs and propose a novel data-agnostic technique, MLP fusion, which leverages clustering characteristics to approximate the NTK. Theoretical derivations are provided to demonstrate how the proposed compression preserves the NTK.
\item We demonstrate that \alg~can be applied to both traditional MLP and MoE modules within PLMs, proving its versatility. 
\item We provide extensive experimental results on PLM SFT for both natural language understanding (NLU) 
and generation (NLG) 
tasks, validating their effectiveness and soundness.
\end{itemize}



\section{Related work}
\label{section:related_work}



There are numerous model compression methods for reducing the size of MLPs in PLMs.
The first line of research, \textbf{knowledge distillation}\cite{sanh2019distilbert,jiao2019tinybert,wang2020minilm}, compresses the pre-trained model and then fine-tunes the compressed model on downstream tasks. Techniques like mean squared error\cite{hinton2015distilling}, optimal transport~\cite{lohit2022model}, and maximum mean discrepancy (MMD)~\cite{huang2017like} are commonly used as distillation loss terms. However, this approach requires loading and executing the large teacher PLM, demanding significant computational resources before fine-tuning.
Another direction involves methods applied \textbf{after the SFT stage} to achieve faster inference. For example, FastBert~\cite{liu2020fastbert} uses a sample-wise adaptive mechanism to adjust inference time, and DeeBERT~\cite{xin2020deebert} accelerates inference by allowing samples to exit earlier. Moefication~\cite{zhang2021moefication} splits MLP modules into sub-networks with a router to select the appropriate sub-network for each input. However, these methods still rely on regular SFT and do not fully utilize PLM knowledge.

In addition to the directions above, a more lightweight efficient fine-tuning paradigm is \textbf{one-shot model compression}. 
As a representative, single-shot pruning methods~\cite{han2015learning, liu2017learning, lee2018snip, wang2020picking, tanaka2020pruning} identify sub-networks at initialization concerning user-specified criteria (e.g.\ weight magnitude or gradient flow) and attain sparsity in model weights. 
The Lottery Ticket Hypothesis (LTH)\cite{frankle2018lottery, chen2020lottery} demonstrates the existence of sparse sub-networks in DNNs and has been applied successfully to PLMs. Recent advancements like NTK-SAP\cite{wang2023ntksap} and PX~\cite{Iurada_2024_CVPR} integrate Neural Tangent Kernel (NTK) theory into pruning, achieving strong performance on architectures like ResNet~\cite{he2015deepresiduallearningimage}. However, pruning primarily reduces theoretical FLOPs, while sparse matrix operations remain inefficient on modern hardware (e.g., GPUs), leading to slower wall-clock training times.
Classical computational techniques like truncated SVD\cite{denton2014exploiting} and randomized sketching\cite{woodruff2014sketching, chen-etal-2022-sketching} are also intuitive for one-shot PLM compression. LoRA~\cite{hu2021loralowrankadaptationlarge} introduces low-rank layers atop original ones, reducing SFT costs by updating only the added parameters. However, LoRA does not decrease inference costs since the final output combines the original and low-rank layers. Detailed trade-offs for LoRA are discussed in \cref{sec:efficiency}.

Specifically for MoE models, various studies have introduced the concept of expert merging \cite{he2023merging,li2023merge,xue2022student,pmlr-v162-liu22k,stoica2024zipitmergingmodelsdifferent} and expert pruning \cite{lu2024expertsequalefficientexpert}, as a method to reduce the number of experts within each layer of the MoE model. However, we note that these methods may not fully leverage the MoE structure, as reducing the number of experts could result in significant information loss.
In \Cref{section:exp} we implement the aforementioned methods as baselines for a comprehensive comparison.

\section{Preliminaries and notations}
\label{section:preliminary}

The notations for MoE layers are introduced in \Cref{sec:pbm_setup}.
We also provide brief preliminaries to NTK in \Cref{sec:NTK_prelim}.

\subsection{Multilayer Perceptron and Mixture-of-Experts Modules} 
\label{sec:pbm_setup}



Across this paper, we denote the input sequence as a feature matrix $\mtx X \in \mb R^{s \times p}$, where $s$ is the sequence length and $p$ is the dimension of the MLP input/output (the dimensions of MLP input and output agree with the embedding dimension in most PLMs).
The neural architecture of interest (a pre-trained transformer) is denoted as $f$, which is different from the scalar loss function $\ell$.
For the simplicity of notation, the output of the whole network $f(\mtx x)$ is assumed to be a \textit{scalar} in this paper, which is the case in regression and binary classification tasks. 
Our derivation however still holds for vector/matrix output if we analyze the output element-wise. 

Specifically, an MLP in the FFN sub-layer of a transformer can be expressed as:
\begin{align}
\label{eqn:mlp_output}
    \mtx H = \sigma(\mtx X \mtx W_1 + \mtx 1 \mtx b_1^\T) \mtx W_2 + \mtx 1 \mtx b_2^\T,
\end{align}
where $\mtx W_1 \in \mb R^{p \times p_I}, \mtx b_1 \in \mb R^{p_I}$ (resp.\ $\mtx W_2 \in \mb R^{p_I \times p}, \mtx b_2 \in \mb R^{p}$) are the weight matrix and the bias term of the first (resp.\ second) linear transform within the FFN sub-layer,
and $\sigma(\cdot)$ is the element-wise activation function. Some other constantly used notations involve the MLP intermediate dimension $p_I$ (the subscript $I$ is short for ``intermediate'').

\begin{figure}[t]
    \centering
    \includegraphics[trim={1cm 2.5cm 0cm 2cm},clip,width=1.0\columnwidth]{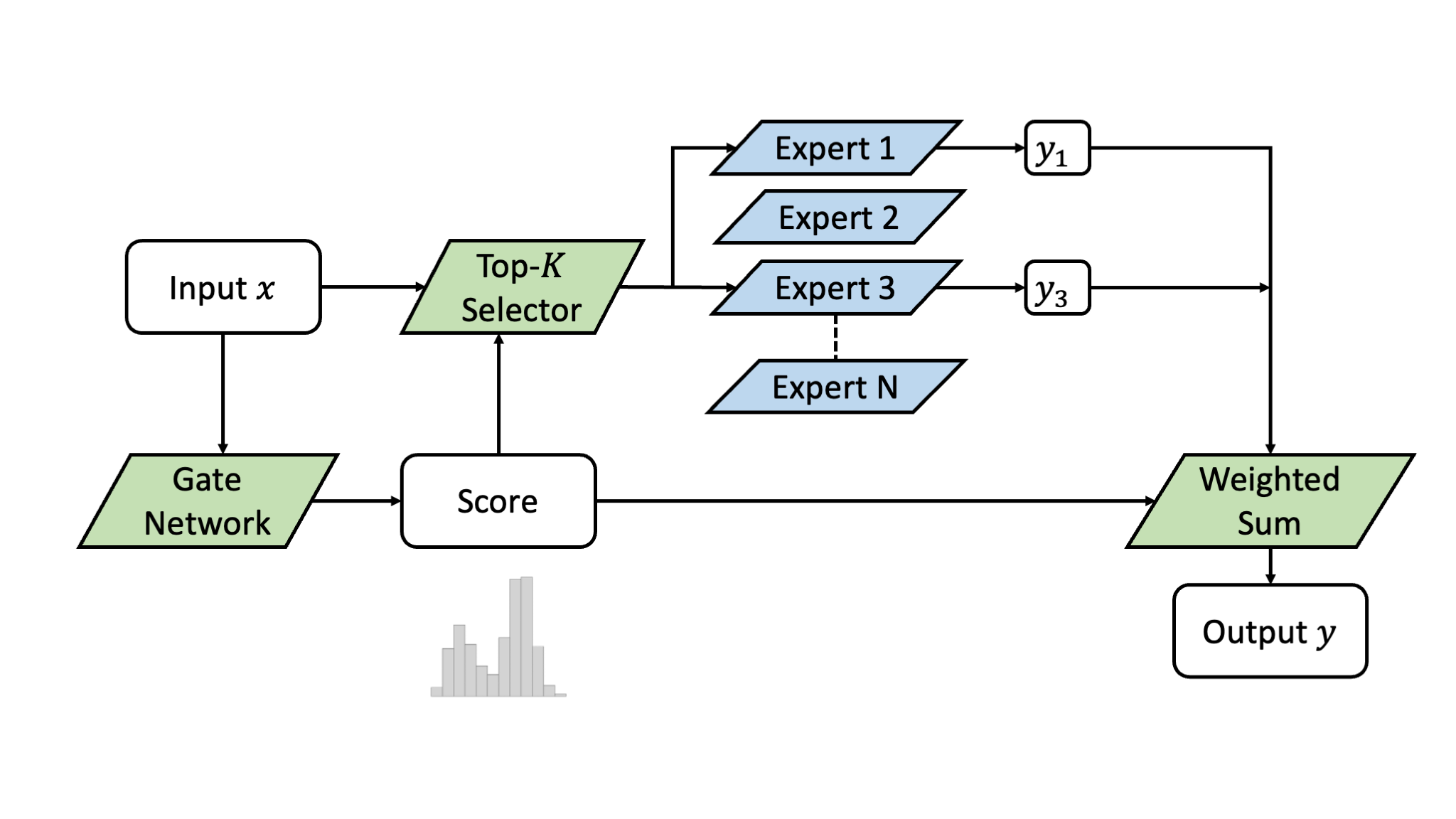}
    \caption{In this illustrative example of MoE layers, the Top-{\em K} Selector, along with the Gate Network--often referred to as the `router'--selects Experts 1 and 3 based on their scores for the given input.
    }
    \label{fig:moe}
\end{figure}

As an improvement over the MLP module above, we also consider the classical mixture-of-experts (MoE) modules, where each expert takes the form of an MLP above. 
We provide the framework illustration of an MoE layer in \cref{fig:moe}.
In more detail, each MoE layer consists of $N$ experts.
The $k$-th expert $E_k$ (a function to transform input vector $\mtx x$ to a new feature) in an FFN sub-layer is an MLP and denoted as:
\begin{align*}
    E_k(\mtx X) = \sigma(\mtx X \mtx W_{1}^{(k)} + \mtx 1 (\mtx b_{1}^{(k)})^\T) \mtx W_{2}^{(k)} + \mtx 1 \mtx b_{2}^{(k)}.
\end{align*}

The output of the MoE layer is thus given by ($\odot$~represents the Hadamard product): 
\begin{align}
\label{eqn:moe_mlp_output}
\mtx H_{m} &=\sum_{k=1}^{N}  [\mtx G(\mtx X)]_{(\cdot,k)}\mtx{1}^\T \odot E_k(\mtx X),
\end{align}
where 
$\mtx G(\mtx X) = \operatorname{Softmax}\left(\operatorname{TopK}\left(\mtx X \mtx W_g \right)\right) \in \mb{R}^{s\times N}$ 
returns the normalized sparse router gating vector of all experts for each token:
$\operatorname{TopK}(\mtx g)$ outputs $\mtx g_i$ when $\mtx g_i$ is within the top-{\em k} values of $\mtx g\in \mathbb{R}^N$, otherwise it returns $-\infty$.
(We slightly abuse the notation in $\operatorname{TopK}\left(\mtx X \mtx W_g \right)$, where $\operatorname{TopK}(\cdot)$ is row-wisely applied to the sequence matrix $\mtx X \mtx W_g$.)
Here, $\mtx W_g \in \mb R^{p \times N}$ represents the linear transform, turning the input token $\mtx x_i$ into the router logit for each expert. 
For example, given a single input vector token $\mtx x_i$, the gate vector $\mtx G(\mtx x)=[0.7,0,0.3,0]$ activates experts 1 and 3 with scores 0.7 and 0.3 (suppose the number of experts $N=4$ and $\operatorname{Top2}$ are taken.) 
The gradients (as well as the NTK) of the weights in an MLP or MoE module are analyzed in \Cref{sec:NTK_form}.

\subsection{Neural Tangent Kernel}
\label{sec:NTK_prelim}

NTK is a powerful theoretical technique to study the gradient descent dynamics of neural networks \cite{jacot2018neural}.
It originated from the research on infinitely wide or ultra-wide neural networks. 
In applying NTK to convolutional neural networks for computer vision tasks, it is noted that NTK can be extended to arbitrary neural architecture $f$ and initialization $\theta_0$, which induces the stochastic gradient descent (SGD) NTK as \cite{arora2019exact}: 
\begin{align}
\label{eqn:ntk}
\dotp{\nabla_{\theta_0} f(x; \theta_0)}{\nabla_{\theta_0} f(z; \theta_0)}.
\end{align}

We note that the NTK above is specific to the 
 SGD~\cite{robbins1951stochastic} optimizer.
As for Adam~\cite{DBLP:journals/corr/KingmaB14}, the most common optimizer for language model fine-tuning, its corresponding NTK in the early stage of training (which is believed to match the nature of the short-period fine-tuning) can be approximated by the so-called Asymmetric SignGD Kernel~\cite{malladi2022kernel}:
\begin{align}
\m K^{(\mathrm{AS})}(x, z) \defeq \dotp{\nabla_{\theta_0} f(x; \theta_0)}{\mathrm{sign}\paren{\nabla_{\theta_0} f(z; \theta_0)}},
\end{align}
we will refer to this kernel by Adam NTK and focus on the analysis thereof in this paper,
considering Adam is the dominant optimizer in language model fine-tuning.

Recent works show directly using the NTK~\cref{eqn:ntk} extracted from a pre-trained model $f(\cdot)$ can obtain decent performance on computer vision tasks~\cite{wei2022more}, 
and in some cases can capture the training dynamics of language model fine-tuning \cite{malladi2022kernel}.
There has already been some discussion on compressing a trained (fine-tuned) network with NTK preserved through pruning and quantization~\cite{gu2022lossless,wang2023ntksap,Iurada_2024_CVPR}.
We will shortly leverage the useful tool to guide the design of our proposed models and serve as a sanity check as well.

\section{MLP fusion with approximate NTK}
\label{section:method}


For the reader's convenience, we first derive the concrete form of the NTK for MLP and MoE modules in \Cref{sec:NTK_form}. Next, \Cref{sec:clustering} outlines the exact form of the proposed clustering-based \alg, followed by verification in \Cref{sec:ntk_verify} to ensure the method meets the previously stated expectations. Finally, \Cref{sec:layer_wise} presents layer-wise tuning, incorporating ideas from further distillation.

\subsection{Preparation: NTK for MLP \& MoE}
\label{sec:NTK_form}

As the gradients w.r.t.\ the model weights are the building blocks of NTK, we provide the expressions of the gradients for MLP and MoE as follows, 
whose calculation is based on $\nabla_{\mtx H} f / \nabla_{\mtx H_m} f_m \in \mb R^{s \times p}$ and the chain rule (similar to $f$, along this paper $f_{m}$ is set as a scalar MoE model).

\textbf{Gradients for MLP modules.} We start with the gradients for a classical MLP module in \Cref{eqn:mlp_output}.
In particular,
\begin{equation}
\label{eqn:grad}
\begin{aligned}
\nabla_{\mtx W_2} f &= \mtx \sigma^\T \nabla_{\mtx H} f, \quad 
\nabla_{\mtx b_2} f = \paren{\nabla_{\mtx H} f}^\T ~ \mtx 1 \\
\nabla_{\mtx W_1} f &= \mtx X^\T \brkt{\paren{\nabla_{\mtx H} f ~ \mtx W_2^\T} \odot \mtx \sigma'} \\
\nabla_{\mtx b_1} f &= \brkt{\paren{\nabla_{\mtx H} f ~ \mtx W_2^\T} \odot \mtx \sigma'}^\T \mtx 1,\\
\end{aligned}
\end{equation}
where 
we by convention abuse the boldfaced notation $\mtx \sigma, \mtx \sigma'$ as a shorthand for $\sigma\paren{\mtx X \mtx W_1 + \mtx 1 \mtx b_1^\T}$ and $\sigma'\paren{\mtx X \mtx W_1 + \mtx 1 \mtx b_1^\T}$ respectively.
The computation of NTK would then be boiled down to the proper inner products of the aforementioned gradient terms. 
It is worth mentioning that the classical model compression technique, pruning, is expected to well approximate the NTK.
Assuming the output of a pruned MLP is close to the original one, the gradient terms will be roughly approximated by the Hadamard product of the mask matrix and the original gradient terms in \Cref{eqn:grad}.

\textbf{Gradients for MoE modules.} 
We first give the matrix form of an MoE module to ease the following gradient derivation;
to the best of our knowledge, we are the first to provide this fundamental form for MoE modules.

For a single token $\mtx{x}_{i} \in \mb{R}^{p}$ (also the $i$-th row in the sequence matrix $\mtx X$), the router matrix is denoted as:
\scalebox{0.77}{
\begin{minipage}{\columnwidth}
\vspace{0.1in}
\begin{align*}
\mtx{R}_i \defeq \operatorname{diag}(\mtx G(\mtx{x}_i)) \otimes \mtx I_{p_{\text {I }}}=\begin{bmatrix}
[\mtx G(\mtx{x}_i)]_1 \cdot \mathbf{I}&   & &  \\
& ... &  & \\
&   & [\mtx G(\mtx{x}_i)]_N \cdot \mtx I
\end{bmatrix}_{N\cdot p_{I} \times N\cdot p_{I}},
\end{align*}
\vspace{0.1in}
\end{minipage}
}
where $\otimes$ is the Kronecker product, $\mtx I$ is the identity matrix.
We recall $\mtx G(\mtx x_i)$ is a sparse vector with length $N$ that contains the gate value. 
(Since $\mtx G(\mtx x_i)$ depends on $\mtx x_i$, $\mtx R_i$ is different for each token and we accordingly add the subscript $i$.)
The weight matrices in the MoE layer are collectively denoted as:
\begin{align*}
\widebar{\mtx W}_1 &=\left(\begin{array}{lll}
\mtx W_1^{(1)}  \cdots  \mtx W_1^{(N)}
\end{array}\right)_{ p \times N\cdot p_{\text {I}}}, \\
\widebar{\mtx W}_{2} &=\left(\begin{array}{lll}
\paren{\mtx W_2^{(1)}}^\T  \cdots \paren{\mtx W_2^{(N)}}^\T
\end{array}\right)^\T_{N\cdot p_{\text {I}} \times p}.
\end{align*}
The $i$-th row of the MoE output can accordingly be expressed as (for simplicity, the bias terms are omitted; also, in practice, most MoE models do not contain them):
\begin{align}
    (\mtx{H}_m)_{i}^{\T} = \sigma\left(\mtx{x}_i^{\T}\widebar{\mtx W}_1 \right) \mtx{R}_i \widebar{\mtx W}_{2},
\label{eqn:moe_y}
\end{align}
in which $\mtx{R}_i$ encapsulates crucial expert knowledge and exhibits high sparsity, as typically only a selected number of experts are activated within each MoE layer. 

The gradients of the MoE layer based on \cref{eqn:moe_y} can be obtained as:
\begin{equation}
\label{eqn:grad_moe}
\begin{aligned}
\nabla_{\mtx{\widebar{W}_{2}}}f_{m} &= \sum_i \mtx R_i^{\T} \cdot \mtx \sigma_i \cdot (\nabla_{\mtx H_m} f_m)_i^{\T},\\
\nabla_{\mtx{\widebar{W}_{1}}}f_{m} &= \sum_i \mtx x_i \brkt{\paren{\mtx R_i \widebar{\mtx W}_{2} (\nabla_{\mtx H_m} f_m)_i} \odot \mtx \sigma'_i}^{\T},
\end{aligned}
\end{equation}
in which $\mtx \sigma_i, \mtx \sigma'_i$ are respectively the $i$-th row of $\mtx \sigma, \mtx \sigma'$.

As a closing remark, we intentionally omit the gradient for the weight matrix $\mtx W_g$ in the router, as in practice it is frozen during the SFT stage; 
this practice comes from the observation that preserving the original PLM's universal world information can enhance their performance~\cite{he2023preserving,mukhoti2023finetuning,dou2023loramoe}. 
Our findings in \Cref{sec:ablation} empirically support this observation, demonstrating freezing $\mtx W_g$ can improve model performance.

\begin{figure}[t]
    \centering
    \includegraphics[width=1.0\columnwidth]{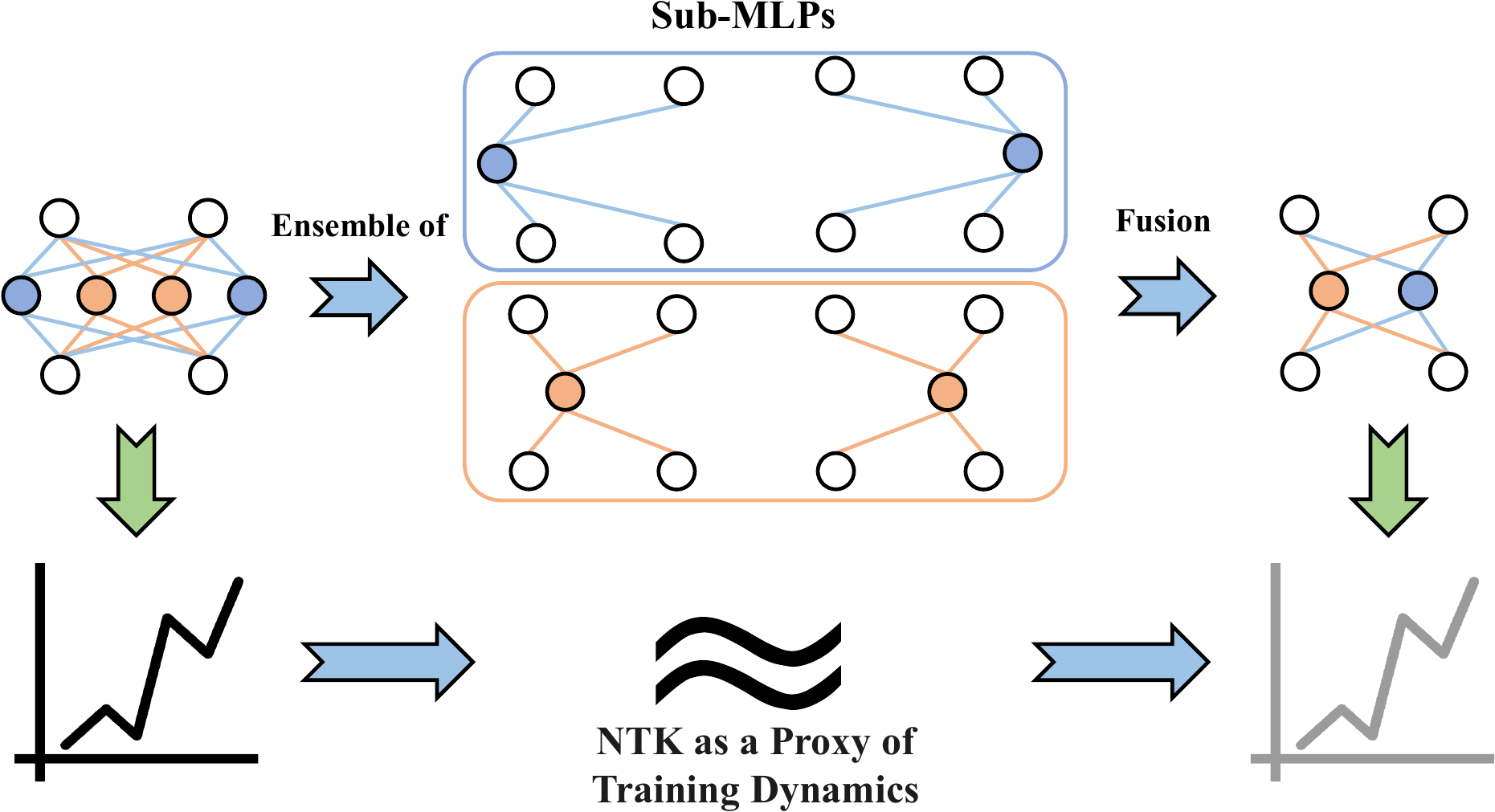}
    \caption{An overview of MLP Fusion. The original MLP in Dense and MoE LLMs can be decomposed as an ensemble of $p_I$ sub-MLPs; through MLP fusion, we cluster the sub-MLPs and re-construct a smaller MLP, which is shown to approximate the NTK of the original MLP and is thus expected to enjoy a similar training dynamics to the full-size PLM.}
    \label{fig:overview}
\end{figure}



\subsection{Methodology: MLP Fusion with Clustering}
\label{sec:clustering}
In this subsection, we primarily develop the proposed method with the goal of ensuring that the new output $\mtx{\wt H}_C$ can effectively approximate the original output $\mtx H$ (a rough approximation analysis is provided in \cref{app:boundmlpoutput} in the supplement), and delay the discussion on NTK approximation to \Cref{sec:NTK_form}. 
We propose \alg~following this intuitive purpose,
through a view that an MLP can be taken as the ensemble of multiple bottleneck-1 sub-MLPs~\cite{chen-etal-2022-inducer, wang2022exploring, yuan2022distributed}.
We rewrite the MLP output in the following form:
\begin{align*}
\mtx H &= \sigma(\mtx X \mtx W_1 + \mtx 1 \mtx b_1^\T) \mtx W_2 + \mtx 1 \mtx b_2^\T \\
&= \sum_{i=1}^{p_I} \left[ \sigma(\mtx X \mtx W_{1, \cdot, i} + \mtx b_{1, i} \mtx 1) \mtx W_{2, i, \cdot}\right] + \mtx 1 \mtx b_2^\T,
\end{align*}
where by convention we represent the $i$-th column (resp.\ row) in the weight matrix $\mtx W_1$ (resp.\ $\mtx W_2$) as $\mtx W_{1, \cdot, i}$ (resp.\ $\mtx W_{2, i, \cdot}$).
The summation implies that it is feasible to approximate MLP via a few bottleneck-1 sub-MLPs (the summands on the right-hand-side above), in a similar way to numerical methods such as importance sampling~\cite{hammersley1954poor} and sketching~\cite{woodruff2014sketching, chen-etal-2022-sketching}.
Considering the existence of the nonlinear activation function $\sigma$, 
we turn to clustering and will demonstrate below how this classical machine learning technique can be used to approximate the above sub-MLP summation.

To obtain the ``embedding'' of the sub-MLPs for clustering, 
we consider the original MLP as $p_I$ supporting points, represented by a design matrix $\mtx W = \brkt{\mtx W_1^\T, \mtx b_1, \mtx W_2} \in \mb R^{p_I \times (2p + 1)}$.
An intuitive idea to compress an MLP is therefore representing the empirical distribution (MLP) by the output centroids of $c$ clusters.
We suggest to use $\mtx w_i = [\mtx W_{1, \cdot, i}^\T, \mtx b_{1, i}, \mtx W_{2, i, \cdot}]^\T$ as the embedding vector for the $i$-th sub-MLP, 
as $\mtx w_i$ can uniquely decide the $i$-th sub-MLP.
Upon the embeddings, Lloyd's algorithm \cite{lloyd1982least} can be directly applied to solve the $k$-means clustering problem, obtain $c$ clusters $\set{\m P_j}_{j=1}^c$,
and return a one-hot clustering matrix~$\mtx C \in \mb R^{c \times p_I}$ with elements $\mtx C_{ji} = 1_{\set{\mtx w_i \in \m P_j}}$.
Normalizing~$\mtx C$ so that the rows sum to $1$, we can construct an averaging matrix $\bar{\mtx C}$ with elements $\bar{\mtx C}_{ji} = \frac{1}{|\m P_j|}_{\set{\mtx w_i \in \m P_j}}$, where $|\m P_j|$ is the number of elements in cluster $j$.
so that $\bar{\mtx C} \mtx W$ will return the desired centroid matrix $\mtx{\wt W} = \brkt{\mtx{\wt W}_1^\T, \mtx{\wt b}_1, \mtx{\wt W}_2} \in \mathbb{R}^{c \times (2 p + 1)}$.
In general, conducting clustering is minimizing the distance from a point to its closest centroid,
which partially explains our intuition that replacing the original sub-MLPs with their corresponding centroids can benefit the compression of MLPs.

\textbf{Relation with Model Fusion}~\cite{singh2020model}.
In principle, we consider the clustering of sub-MLPs shares the same spirit as model fusion, 
which takes a single layer of MLPs as an empirical distribution of the corresponding weights (either $\mtx W_{1, \cdot, i}$'s or $\mtx W_{2, i, \cdot}$'s in our context) and then fuses multiple MLPs into a new one through solving a Wasserstein barycenter problem~\cite{peyre2019computational}.
The clustering procedure is closely related to the problem above, as the output centroids serve as the optimal barycenters when the number of points $\mtx w_i$s assigned to each cluster is fixed.
Due to the connection, we refer to the clustering procedure as MLP fusion in this paper.

\textbf{The derivation for MLP compression}.
We replace each sub-MLP parameter vector $\mtx w_i$ with the corresponding centroid
(equivalently, we replace $\mtx W$ with $\mtx C^\T \mtx{\wt W}$).
The new output can be naturally simplified as: 
\begin{align*}
& \sigma \left(\paren{\mtx X \mtx{\wt W}_1 + \mtx 1 (\mtx{\tilde b}_1)^\T} \mtx C\right) \mtx C^\T \mtx{\wt W}_2 + \mtx 1 \mtx b_2^\T \\
=& \sigma \paren{\mtx X \mtx{\wt W}_1 + \mtx 1 (\mtx{\tilde b}_1)^\T} \paren{\mtx C \mtx C^\T} \mtx{\wt W}_2 + \mtx 1 \mtx b_2^\T,
\end{align*}
where the above equation holds because $\mtx C$ simply ``copies'' the centroids and thus can be taken out of the activation function. 
We will shortly show in \Cref{sec:ntk_verify} that the computational properties of the one-hot clustering matrix $\mtx C$ are indeed critical for Adam NTK approximation.


After being taken out of the activation function, $\mtx C$ is then allowed to be combined with $\mtx C^\T$ to form the scaling matrix referred to as $\mtx P=\mtx C \mtx C^\T$, which is a $c \times c$ diagonal fixed matrix that greatly reduces the computation compared with the original equation.
Note the architecture of the final MLP has not yet been specified, since there are different ways to address the scaling matrix $\mtx P$: it can
\begin{itemize}
    \item either be incorporated into $\mtx{\wt W}_2$, or
    \item stands alone as a constant scaling matrix.
\end{itemize}

It is worth noting that both strategies behave identically during forward propagation; however, during backward propagation, the gradient of the second method is multiplied by the scaling matrix $\mtx P$.
In the final version of \alg, we adopt the form that uses the stand-alone scaling matrix $\mtx P$, and refer to the variant that incorporates $\mtx P$ into $\wt{\mtx W}_2$ as ``clustering''.

\textbf{The derivation for MoE module compression}. 
As an MoE module is composed of multiple MLPs, the proposed \alg~can be respectively applied to the experts therein, and 
the derivation will be similar; we defer the presentation to the next subsection.


\subsection{NTK Approximation}
\label{sec:ntk_verify}

In addition to the aforementioned goal of output approximation, we further suggest a compression method for \emph{fine-tuning} is supposed to preserve the NTK of the original model so that the training dynamics thereof can be preserved as well.

To this end, we revisit the scaling matrix $\mtx P$ and render it ``stand-alone'' (the second choice in the previous subsection);
with this specific design, \alg~is able to approximate the NTK and its form is formally given as:
\begin{align}
\label{eqn:clustering_mlp}
\mtx{\wt H}_C \defeq \sigma \paren{\mtx X \mtx{\wt W}_1 + \mtx 1 \mtx{\wt b}_1^\T} \mtx P \mtx{\wt W}_2 + \mtx 1 \mtx b_2^\T
\end{align}
where we choose $\mtx{\wt W}_1 \defeq \mtx{W}_1 \bar{\mtx C}^\T$, $\mtx{\wt b}_1 \defeq \bar{\mtx C} \mtx{b}_1$, $\mtx{\wt W}_2 \defeq \bar{\mtx C} \mtx{W}_2$ as the new parameters for the compressed MLP, and $\mtx P$ is designed to stand alone as a constant scaling matrix. 
We note that $\mathbf P_{ii}=(\mathbf{CC}^\T)_{ii} = \sum_q \mathbf C_{iq}^2 = \sum_q \mathbf C_{iq}$ ($\mathbf P_{ij}=(\mathbf{CC}^\T)_{ij} = 0$ for $i \neq j$) represents the number of points in cluster $i$. 
From an intuitive perspective, the backward process of multiplying the gradient by the scaling matrix $\mathbf P$ can be seen as assigning different learning rates to different clusters. 
This means that larger clusters are given a larger learning rate. 

As for the \textbf{MoE modules}, directly applying \alg~to each expert will give:
\begin{align}
\label{eqn:clustering_moe}
    (\wt{\mtx{H}}_m)_{i}^{\T} = \sigma\left(\mtx x_i^{\T} \widetilde{\mtx W}_1 \right)\mtx P^{(m)}_i \widetilde{\mtx W}_2,
\end{align}
where $\mtx P^{(m)}_i = \mtx{C}^{(m)} \mathbf{R}_i (\mtx{C}^{(m)})^{\top}$ and the concatenated clustering matrix $\mtx C^{(m)}$ is denoted as:
\begin{align*}
\mathbf{C}^{(m)} = 
    \begin{bmatrix}
	  \mathbf{C}_{1} &    &  \\
	  & ... &   \\
	  &    & \mathbf{C}_{N}
    \end{bmatrix}_{N\cdot c \times N\cdot \pI},
\end{align*}
and $\mathbf{C}_{k}$ is the corresponding clustering matrix specific to expert $k$;
to ease the notations, we also reload the compressed weight matrices $\widetilde{\mtx W}_1, \widetilde{\mtx W}_2$ as
\begin{align*}
\widetilde{\mtx W}_1&=\left(\begin{array}{lll}
\widetilde{\mtx W}_1^{(1)}  \cdots  \widetilde{\mtx W}_1^{(N)}
\end{array}\right), \\
\widetilde{\mtx W}_2&=\left(\begin{array}{lll}
\paren{\widetilde{\mtx W}_2^{(1)}}^\T  \cdots  \paren{\widetilde{\mtx W}_2^{(N)}}^\T
\end{array}\right)^\T,
\end{align*}
where $\widetilde{\mtx W}_1^{(k)} = \mtx{{W}}^{(k)}_1 \bar{\mtx{C}}_k^{\T}, \widetilde{\mtx W}_2^{(k)} = \bar{\mtx{C}}_k\mtx{{W}}^{(k)}_2, \forall k \in [N]$, and similar to $\bar{\mtx{C}}$, $\bar{\mtx{C}}_k$ is the normalized version of $\mtx C_k$. 
A closing remark is that the compression version in \Cref{eqn:clustering_moe} enjoys the same form as the original MoE output in \Cref{eqn:moe_y}, and thus we can similarly derive the gradient for $\widetilde{\mtx W}_1, \widetilde{\mtx W}_2$ in the MoE modules as in \Cref{eqn:grad_moe}.

We will then show how the specific MLP \eqref{eqn:clustering_mlp} can serve to approximate the NTK of the original MLP \eqref{eqn:mlp_output} (the derivation for the MoE module~\eqref{eqn:moe_mlp_output} can be found in the supplementary material).
This expectation implies the following requirements:
first, the new output $\mtx{\wt H}_C$ (resp. $\wt{\mtx{H}}_m$)  is supposed to approximate the original output $\mtx H$ (resp. ${\mtx{H}}_m$),
which we have heuristically shown in the previous discussion;
second, the hidden representation $\mtx \sigma$ and the related composition term $\paren{\nabla_{\mtx H} f ~ \mtx W_2^\T} \odot \mtx \sigma'$ should also be preserved. 


This subsection will thus be devoted to verifying that the proposed method can induce an Adam NTK close to the original one.
The key step is to show the inner product of the four gradient terms in \Cref{eqn:grad} will approximately remain.
We prepare some additional notations to ease the following discussions and let the compressed neural model be equipped with the compressed MLP module as $f_c$.
The input token sequence is denoted as $\mtx X$ or $\mtx Z$, respectively.

We first make an assumption that the clustering can capture the MLP empirical distribution, so that, to some sense, $\mtx C \mtx{\wt W} \approx \mtx W$ and $\mtx{\wt H}_C \approx \mtx H$ as derived in \Cref{sec:clustering}. 
This assumption implies that $\nabla_{\mtx H} f$ can be preserved by $\nabla_{\mtx{\wt H}_C} f_c$, since they depend on $\mtx H$ / $\mtx{\wt H}_C$ in the same way.
We can automatically obtain $\dotp{\nabla_{\mtx b_2} f(\mtx X)}{\mathrm{sign}\paren{\nabla_{\mtx b_2} f(\mtx Z)}} \approx \dotp{\nabla_{\mtx{b}_2} f_c(\mtx X)}{\mathrm{sign}\paren{\nabla_{\mtx{b}_2} f_c(\mtx Z)}}$.

We then analyze the term $\dotp{\nabla_{\mtx W_2} f(\mtx X)}{\mathrm{sign}\paren{\nabla_{\mtx W_2} f(\mtx Z)}}$,
where the notation $\dotp{\cdot}{\cdot}$ is reloaded as the matrix inner product $\dotp{\mtx X}{\mtx Z} \defeq \Tr\paren{\mtx X^\T \mtx Z}$.
The term equals\footnote{$\mtx \sigma_x, \mtx \sigma_z$ are the shorthand for $\sigma\paren{\mtx X \mtx W_1 + \mtx 1 \mtx b_1^\T}$ and $\sigma\paren{\mtx Z \mtx W_1 + \mtx 1 \mtx b_1^\T}$ respectively.
Similarly, we define $\mtx{\tilde\sigma}_x \defeq \sigma\paren{\mtx X \mtx{\wt W}_1 + \mtx 1 \mtx{\tilde b}_1^\T}$ and $\mtx{\tilde\sigma}_z \defeq \sigma\paren{\mtx Z \mtx{\wt W}_1 + \mtx 1 \mtx{\tilde b}_1^\T}$ for $f_c$.}
$$\Tr\brkt{\paren{\nabla_{\mtx H} f(\mtx X)}^\T \mtx \sigma_x \cdot
\mathrm{sign}\paren{\mtx \sigma_z^\T \nabla_{\mtx H} f(\mtx Z)}},$$
and can be shown to approach 
$$\dotp{\nabla_{\mtx{\wt W}_2} f_c(\mtx X)}{\mathrm{sign}\paren{\nabla_{\mtx{\wt W}_2} f_c(\mtx Z)}}.$$
Concretely, we re-utilize the deduction $\nabla_{\mtx H} f \approx \nabla_{\mtx{\wt H}_C} f_c$ to make it sufficient to study whether $\paren{\mtx{\tilde\sigma}_x \mtx P} \cdot \mathrm{sign}\paren{\mtx P \mtx{\tilde\sigma}_z^\T}$ can approximate its counterpart $\mtx \sigma_x \cdot \mathrm{sign}\paren{\mtx \sigma_z^\T}$.

Analogous analyses of the matrix product 
$$\brkt{\paren{\nabla_{\mtx H} f(\mtx X) \mtx W_2^\T} \odot \mtx \sigma'_x} \cdot \brkt{\paren{\nabla_{\mtx H} f(\mtx Z) \mtx W_2^\T} \odot \mtx \sigma'_z}^\T$$ 
can also be performed for the other two terms 
\begin{align*}
&\dotp{\nabla_{\mtx W_1} f(\mtx X)}{\mathrm{sign}\paren{\nabla_{\mtx W_1} f(\mtx Z)}} \quad \text{and} \\
&\dotp{\nabla_{\mtx b_1} f(\mtx X)}{\mathrm{sign}\paren{\nabla_{\mtx b_1} f(\mtx Z)}}.
\end{align*}
Due to space constraints, derivations for the two terms and the MoE modules are provided in the supplementary Appendix B-C \& B-D.
Notably, the element-wise sign function in the Adam optimizer plays a key role in NTK preservation. For regular SGD, additional conditions and modifications are required. Readers can find more details in  Appendix B-E.

\subsection{Layer-wise Task-specific Tuning}
\label{sec:layer_wise}

In the previous section,
\alg~manages to exploit the potentials of the pre-trained models in a one-shot and task-agnostic manner,
where we retain the training dynamics of neural networks through NTK preservation. 
To more effectively acquire the knowledge within each task, we can leverage the idea from distillation and intuitively design a layer-wise (and thus lightweight) task-specific tuning module, which further tunes the fused MLP with task-specific unsupervised training data. 
Compared to classical distillation, the layer-wise tuning lasts a shorter period (only $1$ epoch in our experiments in \Cref{section:exp},) and we only updates the weights in the fused MLP.

To be specific, we set the tuning loss as the mean squared error (MSE) between the layer output $\mtx H^l_t$ in the teacher model and the layer output $\mtx H^l$ in the student model for layer $l$ of the PLM. 
The tuning loss is then computed as:
\begin{equation}
    \ell^{\text{tune}} = \sum_{l=1}^{L} \text{MSE}(\mtx H^l_t, \mtx H^l)
\end{equation}
where $L$ is the number of layers in the PLM and MSE denotes the mean squared error. 

\section{Numerical results}
\label{section:exp}


In this section, we present the numerical results of MLP Fusion compared to representative baselines on both NLU and NLG tasks. We start with the experimental setup and an initial evaluation of the approximation error, followed by an ablation study and efficiency evaluation. Detailed experimental information is included in the supplementary material.


\subsection{Experiment Setup}

\subsubsection{Backbone Models} 
As our experiments cover both NLU and NLG tasks, we leverage different model types tailored to each. Specifically, we utilize RoBERTa~\cite{liu2019roberta}, an encoder-only architecture, for the NLU tasks, and GPT-2~\cite{radford2019language}, a decoder-only architecture, for the NLG tasks. For MoE models, we employ the Switch Transformer~\cite{fedus2022switch}, an encoder-decoder model with 16 experts in each MoE layer.

\subsubsection{Baseline Methods}
We mainly compare our method with one-shot compression methods:
regular fine-tuning \cite{howard2018universal}, truncated SVD~\cite{denton2014exploiting}, Pruning (single-shot unstructured pruning)~\cite{han2015learning, lee2018snip, wang2020picking, tanaka2020pruning}, LTH (Lottery Ticket Hypothesis)~\cite{frankle2018lottery,chen2020lottery},  GEM-MINER~\cite{osti_10395564}, Moefication~\cite{zhang2021moefication}. We also compare the structured pruning method FLOP (Factorized Low-rank Pruning)~\cite{wang-etal-2020-structured}. 
In addition, we list the performance of DistilRoBERTa~\cite{Sanh2019DistilBERTAD} as a reference for NLU tasks. 

By default, all the baseline methods are set to reduce the MLP intermediate size from 3076 to 768 or a comparable number of parameters.
Specifically, (\rom{1})~truncated SVD 
keeps only the $t$ largest singular values and the associated singular vectors.
(\rom{2})~Unstructured pruning globally removes a certain ratio of connections by exploring the weight magnitude and gradient. 
Concretely, we mask 75\% of the connections to match the compression rate (25\%). As a special case of pruning, (\rom{3})~Lottery tickets hypothesis (LTH) \cite{frankle2018lottery,chen2020lottery}, demonstrates the existence of sparse subnetworks in DNNs, which performs iterative sparsification during tuning to find the matching network. We iteratively prune the MLP for one epoch to make it computationally equivalent to the layer-wise task-specific tuning module (introduced in \Cref{sec:layer_wise}). 
The network mask ratio is also 75\% as pruning. 
(\rom{4})~Moefication splits the MLP modules of a PLM into several sub-networks and designs the additional route mechanism to decide the corresponding sub-network for each input. 
Here we split the original MLP into 4 sub-networks to match the network compression ratio.

Furthermore, we introduce three machine learning techniques, randomized sketching, MMD approximation, and regular clustering, as extra strong baselines. 
We demonstrate as follows how to apply them to MLP compression.

\noindent\textbf{Sketching the Weight Matrices.}
The idea of Sketching is to reduce the size of $\mtx W_1, \mtx W_2$ by multiplying a matrix $\mtx S$:
\begin{align*}
    \mtx{\wt H}_S = \sigma(\mtx X \mtx{W_1} \mtx S + \mtx 1 \mtx b_1^\T \mtx S) \mtx S^\T \mtx W_2 + \mtx 1 \mtx b_2^\T,
\end{align*}
where $\mtx S$ can be a Gaussian Sketching Matrix, which applies Johnson–Lindenstrauss transform \cite{ailon2009fast} to the weight matrices $\mtx W_1, \mtx W_2$.
We expect Sketching can more or less preserve the information within the PLM.

\noindent\textbf{Compression via Minimizing MMD.}
As described in \cref{sec:clustering}, we propose viewing the MLP as an empirical distribution of sub-MLPs. Intuitively, the MLP can be compressed by minimizing the MMD distance between the original and compressed MLP distributions. This approach produces a compressed model with fewer support points while preserving key properties (details are provided in Appendix~\ref{app:mmd} in the supplement).

\noindent\textbf{Clustering without NTK approximation.}
Moreover, to clearly ablate the effect of NTK approximation, we implement a clustering-based method where the fused weights $\mtx{\wt W}_1$, $\mtx{\tilde b}_1$, and $\mtx{\wt W}_2$ are replaced as follows:

$\mtx{\wt W}_1 = \mtx{W}_1 \bar{\mtx C}^\T \mtx P^\frac12,$
$\mtx{\tilde b}_1 = \bar{\mtx C} \mtx{b}_1 \mtx P^\frac12,$
$\mtx{\wt W}_2 = \mtx P^\frac12 \bar{\mtx C} \mtx{W}_2.$

Here, $\mtx P = \mtx{CC}^\T$ is a diagonal matrix. The corresponding MLP is then defined as:
$$\sigma \paren{\mtx X \mtx{\wt W}_1 + \mtx 1 \mtx{\tilde b}_1^\T} \mtx{\wt W}_2 + \mtx 1 \mtx b_2^\T,$$
which enjoys the same architecture as ``Sketching'' and ``MMD''.

For the layer-wise task-specific tuning, we adopt the original RoBERTa \cite{liu2019roberta}/Switch Transformer~\cite{fedus2022switch} (GPT-2 \cite{radford2019language} for language generation) as the teacher and the language model with fused MLP as the student, on the two NLU tasks.
The tuning only lasts for $1$ epoch so as to match the computational cost of the pre-processing in LTH.

\begin{table}[t]
\centering
\caption{Approximation error of each baseline method on SST2 validation set with RoBERTa as the backbone.}
\label{table:ntk}
\resizebox{\columnwidth}{!}{
\begin{tabular}{lcc}
\toprule
\multirow{2}{*}{} & \multicolumn{2}{c}{Approximation Error} \\
            & \multicolumn{1}{c}{Output}      & \multicolumn{1}{c}{NTK}                  \\
\midrule
Sketch       & 24.48$\pm$0.61 & 242757.23$\pm$42629.38 \\ 
MMD          & 8.92$\pm$0.22  & 7620.49$\pm$527.86     \\ 
SVD          & 5.89$\pm$0.00  & 4423.38$\pm$108.89     \\ 
Pruning        & 5.18$\pm$0.23  & 6623.20$\pm$463.72     \\ 
LTH          & 5.10$\pm$0.18  & 6628.73$\pm$462.03     \\ 
Clustering      & \textbf{4.83$\pm$0.02}  & 7030.91$\pm$561.52     \\
\midrule
MLP Fusion (Ours) & \textbf{4.83$\pm$0.02}  & \textbf{2826.59$\pm$155.06}     \\
\bottomrule
\end{tabular}
}
\vspace{-0.5cm}
\end{table}

\subsection{Preliminary Evaluation of Approximation Error}

As a sanity check, we first perform the preliminary evaluation of NTK approximation error for each applicable method.
We examine the output and the induced NTK of the first-layer MLP on the validation set of SST2 with RoBERTa-base \cite{liu2019roberta} compressed by different baseline approaches.
The results, averaged over three runs, are summarized in Table \ref{table:ntk}. Note that DistilRoBERTa and Moefication were excluded from this analysis due to their focus on different objectives.
Specifically, SVD is a deterministic method and therefore gives $0$ standard deviation.
We come up with the following observations: 
(\rom{1})~Most of the listed methods can well approximate the MLP output with a small output distance between the original RoBERTa and the compressed model. 
(\rom{2})~MLP Fusion achieves the smallest NTK approximation error among all methods, defined as the $l_2$ difference between the NTK kernel matrix computed on the evaluation samples before and after compression. 
This experiment verifies our proposal that MLP Fusion best preserves the training dynamics of the original PLM.

\subsection{Experiments on Natural Language Understanding}
\label{subsection:nlu}
We provide extensive experimental comparisons based on RoBERTa as the PLM with a set of representative baselines on natural language understanding benchmarks SST2, MNLI, which can be found in Table \ref{table:nlu}, while additional test results are provided in 
the supplementary material. Furthermore, we present performance comparisons among various methods after task-specific fine-tuning and two additional baselines that try to maintain MLP output and NTK in the Appendix.

\begin{table}[t]
\centering
\caption{Performance (ACC) on SST2 and MNLI validation sets with RoBERTa as the PLM. \textbf{Bold} indicates the best score for each metric, while \textbf{underlined} values represent the second-best.
}
\label{table:nlu}
\addtolength{\tabcolsep}{-2pt}
\scalebox{1}{
\begin{tabular}{lcc}
\toprule 
        & SST2          & MNLI         \\
\midrule
RoBERTa       & 94.61$\pm$0.09  & 87.34$\pm$0.28        \\
DistilRoBERTa & 92.50$\pm$0.12   & 84.03$\pm$0.18  \\ 
\midrule
Sketch        & 91.90$\pm$0.14  & 83.30$\pm$0.11        \\
MMD           & 92.54$\pm$0.41 & 84.20$\pm$0.24         \\
SVD           & 92.55$\pm$0.24 & 85.23$\pm$0.04        \\ 
FLOP              & 92.12$\pm$0.19     & 84.05$\pm$0.21     \\
Pruning        & 92.78$\pm$0.17 & 85.82$\pm$0.12        \\
LTH           & 92.91$\pm$0.15  & 85.96$\pm$0.10        \\
GEM-MINER         & 92.89$\pm$0.16     & 85.51$\pm$0.11     \\
Moefication   & 92.19$\pm$0.20  & 84.83$\pm$0.27 \\
Clustering       & 93.01$\pm$0.17 & 85.75$\pm$0.04         \\
\midrule
MLP Fusion (Ours)    & \underline{93.23$\pm$0.23}  & \underline{86.10$\pm$0.06} \\
+Task-specific Tuning      & \textbf{93.79$\pm$0.07  }            & \textbf{86.32$\pm$0.06}           \\ 
\bottomrule
\end{tabular}}
\vspace{-0.3cm}
\end{table}
DistilRoBERTa, a lightweight version of RoBERTa with the same training process, served as a baseline for comparison. For all compressed methods, we reduced the intermediate size to 25\% ($3072 \rightarrow 768$), aligning with the MLP input size. Pruning/LTH involved masking 75\% of MLP connections, while Moefication divided the MLP into four experts to reduce model size. All reductions were applied to the last 8 layers of the PLM to ensure fairness.

From the table, we have the following findings: 
(\rom{1})~MLP Fusion outperforms all the baselines, which demonstrates the effectiveness of our proposed approach. 
(\rom{2})~Without NTK approximation, there is an obvious reduction in the performance of Clustering, which verifies the necessity thereof. 
(\rom{3})~Layer-wise task-specific tuning further enhances the performance of MLP Fusion by incorporating task-specific knowledge. 
It is worth noting that LTH and GEM-MINER also require pre-processing when masking connections; after making them computationally comparable to \alg, 
the superior performance of our method more clearly validates its edges. 






For the MoE model Switch Transformer, experiments were conducted under similar conditions (c.f.\ the results in \cref{tab:switch}). Unlike RoBERTa, a distilled version of Switch Transformer was unavailable. In addition to selecting the best-performing baseline from \cref{table:nlu}, we included M-SMoE \cite{li2023merge}, a MoE-specific SFT method, and NTK-SAP \cite{wang2023ntksap}, which combines NTK with pruning. 
Similar conclusions can be drawn that 
(\rom{1})~Clustering's performance falls behind compared to \alg, which generally outperforms the baseline methods. Furthermore, after additional distillation, the performance improves even more.
(\rom{2})~It is worth noting that even though NTK-SAP produces a commendable result, the actual speed-up is limited, as further discussed in \cref{sec:efficiency}.

\subsection{Experiments on Natural Language Generation}

In this part, we investigate the effectiveness of the proposed MLP Fusion by evaluating on the natural language generation benchmark WebNLG with a set of one-shot comparable baselines. The results are reported in Table \ref{table:nlg}. The compressing/pruning setup is the same as the NLU evaluation shown in Section \ref{subsection:nlu}.
Our proposed method generally preserves the performance of the naive fine-tuning approach most effectively. MLP Fusion achieves an average accuracy improvement of about 1$\%$ compared to the baselines. Among the baselines, the pruning method stands out due to its preservation of the original MLP weight matrix size. However, the lack of robust support for sparse matrix multiplication on modern GPU hardware limits the actual efficiency gains, as detailed in \cref{sec:efficiency}.

\begin{table*}[t]
\centering
\caption{
Evaluation results of Switch Transformer on four GLUE NLU tasks (measured in accuracy). \textbf{Bold} indicates the best score for each metric, while \textbf{underlined} values represent the second-best.
}
\label{tab:switch}
\begin{tabular}{lccccc}
\toprule
&SST2 & MRPC & CoLA & MNLI \\
\midrule
Switch Transformer &95.60$\pm$0.08  & 91.17$\pm$0.12  & 83.48$\pm$0.17 & 88.67$\pm$0.06 \\
\midrule


Sketch        & 91.25$\pm$0.23   & 69.69$\pm$0.38  & 69.12$\pm$0.00 & 83.36$\pm$0.28      \\
MMD           & 94.19$\pm$0.04 & 87.91$\pm$0.29 & 81.50$\pm$0.03 &   87.25$\pm$0.02    \\

Pruning & 93.81$\pm$0.01  & 88.48$\pm$0.22 & 82.20$\pm$0.13  & 87.20$\pm$0.08  \\
SVD & 94.53$\pm$0.08  & 89.05$\pm$0.05 & 82.36$\pm$0.03  & 87.85$\pm$0.03  \\
M-SMoE & 94.84$\pm$0.03 & 89.05$\pm$0.06 & 82.36$\pm$0.03  & 87.51$\pm$0.03  \\
NTK-SAP & 94.88$\pm$0.02  & 89.30$\pm$0.12  & \underline{82.61$\pm$0.05}  & 87.85$\pm$0.03 \\
Clustering & 94.76$\pm$0.02 &89.38$\pm$0.20 &82.13$\pm$0.02 & 87.50$\pm$0.02  \\

\midrule
\alg & \underline{95.03$\pm$0.15} & \underline{89.87$\pm$0.13} & 82.39$\pm$0.02 & \underline{87.89$\pm$0.01}  \\
+Task-specific Tuning &\textbf{95.15$\pm$0.08}&\textbf{89.95$\pm$0.08}&\textbf{82.65$\pm$0.01}&\textbf{87.91$\pm$0.02}\\

\bottomrule
\end{tabular}
\end{table*}

\begin{table*}[htbp]
\centering
\caption
{Performance (\%) of baseline methods on WebNLG with GPT-2 as the PLM. \textbf{Bold} indicates the best score for each metric, with a down-arrow denoting that lower values are better. \textbf{Underlined} values represent the second-best results.
}

\centering
\label{table:nlg}
\begin{tabular}{lccccccccccc}
\toprule
\multirow{3}{*}{} & \multicolumn{9}{c}{WebNLG}                                                                                                                                                                                                                                             \\
                  & \multicolumn{3}{c}{BLEU}                                                          & \multicolumn{3}{c}{MET}                                                        & \multicolumn{3}{c}{TER $\downarrow$}                                                                           \\
                  & S                         & U                         & A                         & S                        & U                        & A                        & S                        & U                        & \multicolumn{1}{c}{A}                                            \\ 
\midrule
GPT-2       & 57.93                     & 22.55                     & 42.17                     & 0.42                     & 0.25                     & 0.34                     & 0.39                     & 0.76                     & 0.56                                      \\ 
\midrule
Sketch            & 43.16                     & 8.07                      & 27.28                     & 0.32                     & 0.13                     & 0.23                     & 0.54                     & 0.95                     & 0.73                                           \\
MMD               & 56.73                     & 19.90                     & 40.17                     & 0.41                     & 0.23                     & 0.33                     & 0.41                     & 0.79                     & 0.59                                           \\
Pruning             & 55.21                     & \textbf{21.80}                     & 40.55                     & 0.40                     & 0.25                     & 0.33                     & 0.42                     & \textbf{0.76}                     & \yc{0.57}                                         \\
Clustering           & 54.75                     & 20.63                     & 39.81                     & 0.40                     & 0.25                     & 0.33                     & 0.43                     & 0.77                     & 0.59                                          \\ 
\midrule
MLP Fusion (Ours)       & \textbf{57.12}                     & 21.03                     & \underline{40.79}                     & \underline{0.41}                     & \underline{0.25}                     & \underline{0.33}                     & \underline{0.41}                     & 0.78                     & \underline{0.58}                                          \\
+Task-specific Tuning        & \underline{56.75}  & \underline{21.41}  & \textbf{41.04}  & \textbf{0.42}   & \textbf{0.25}   & \textbf{0.33}  & \textbf{0.40}   & \underline{0.77}   & \textbf{0.57}      \\ 
\bottomrule
\end{tabular}
    \vskip 0.05in
    \raggedright
    \small\textsuperscript{a} The letters S, U, and A in the WebNLG metric denote SEEN, UNSEEN, and ALL; instances under the SEEN categories are used for training; instances under the UNSEEN categories are used for testing; ALL has all the instances in it.
\end{table*}




\begin{figure}[t]
    \centering
    \includegraphics[width=1.\columnwidth]{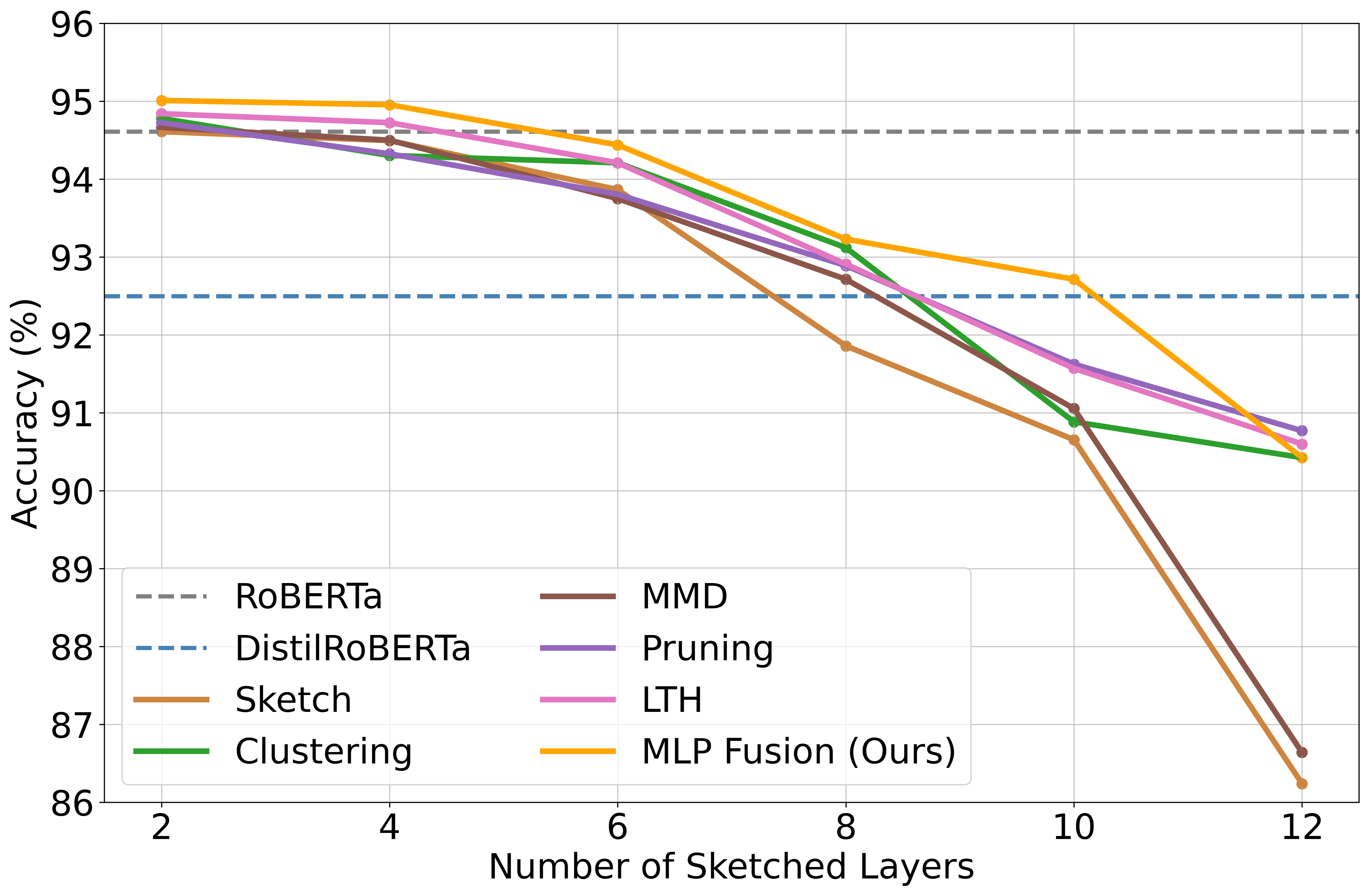}
    \caption{Accuracy of each baseline method with respect to various numbers of sketched layers on the SST2 data set.}
    \label{fig:sketch_layer}
\end{figure}

\begin{figure}[t]
    \centering
    \includegraphics[width=1\columnwidth]{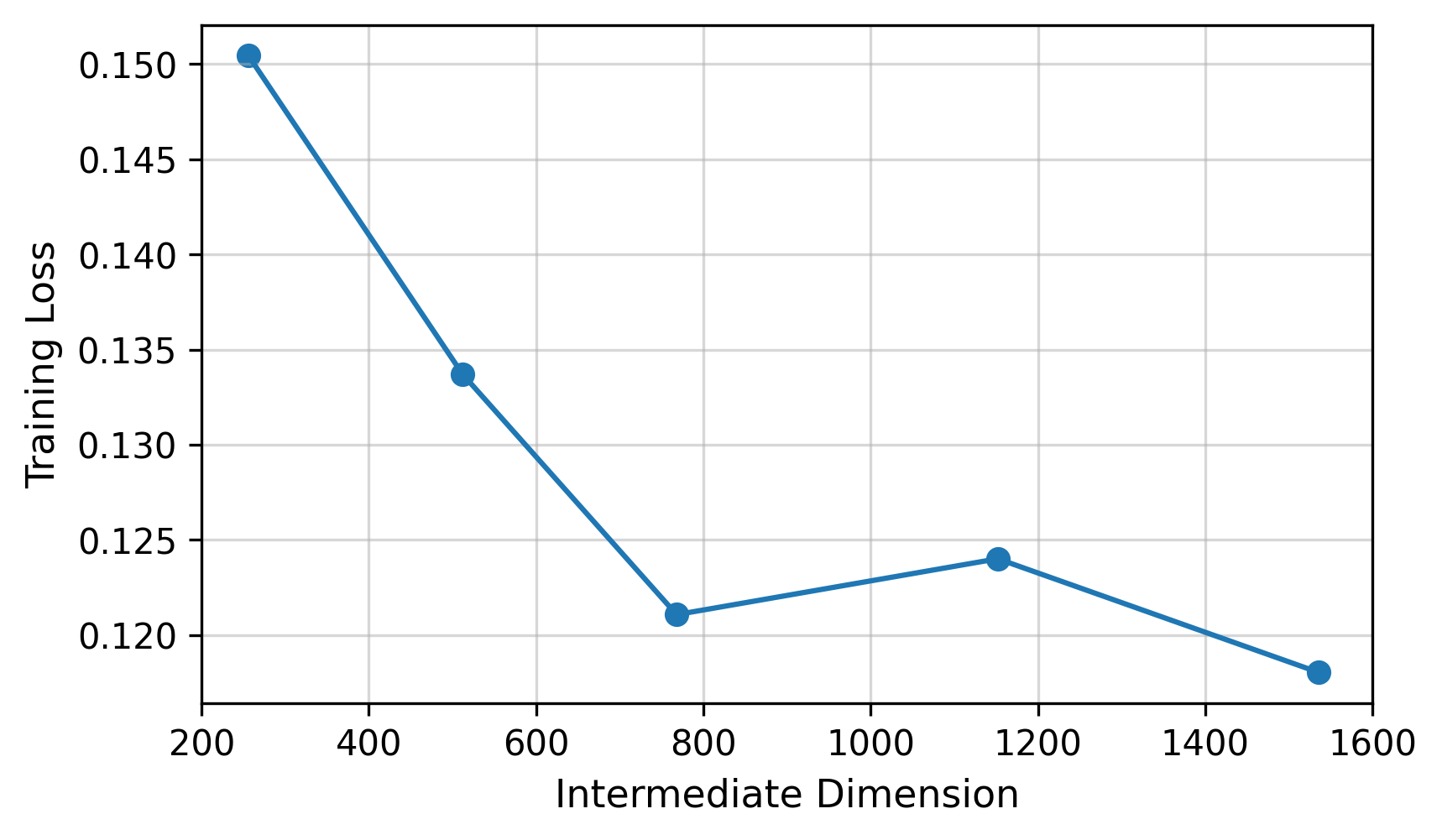}
    \caption{Training loss (dynamic) of our proposed MLP Fusion with respect to various intermediate dimensions on the SST2 data set.}
    \label{fig:loss_dim}
\end{figure}

\subsection{Ablation Studies}

\label{sec:ablation}

\textbf{Impact of Sketch Layers.}
To analyze the effect of sketching layers in PLMs, we extended our experiments beyond sketching the last 8 layers on SST-2. Figure \ref{fig:sketch_layer} provides insights, with dashed lines representing the performance of RoBERTa and DistilRoBERTa. The horizontal axis value "2" corresponds to sketching the last 2 layers.
MLP Fusion consistently achieves comparable or superior performance to the baselines, often outperforming raw RoBERTa when fewer than 6 layers are sketched. This demonstrates the potential of MLP Fusion to reduce redundancy in neural networks, effectively offering a "free lunch" in PLM fine-tuning. Moreover, MLP Fusion consistently surpasses DistilRoBERTa as long as fewer than 10 layers are sketched.
However, sketching all 12 layers significantly reduces performance, indicating that the bottom layers of PLMs retain critical semantic information. This observation aligns with findings in \cite{zhang2020revisiting}.


\textbf{The Choice of the Intermediate Dimension in MLP Fusion.}
In our initial experiments, we set the intermediate dimension to $768$, matching the MLP input size. To explore its impact, we analyzed the training dynamics and testing performance of MLP Fusion across various intermediate sizes. As shown in Figure \ref{fig:loss_dim}, the training loss remains stable when the intermediate dimension is $768$ or larger. However, reducing the dimension below $768$ results in a sharp increase in loss due to the MLP being rank-deficient, which adversely affects performance. Testing results, presented in Table \ref{table:diff_inter_perf}, exhibit a similar trend, underscoring the importance of maintaining sufficient intermediate dimensionality.


\begin{table}[t]
\centering
\caption{Performance of proposed method and the baseline method at different levels of compression.}
\label{table:diff_inter_perf}
\begin{tabular}{lcc}
\toprule
Intermediate Dimension & Sketch & MLP Fusion \\
\midrule
256                        & 88.19  & 90.71      \\ 
768                        & 91.90  & 93.23      \\ 
1536                       & 92.09  & 93.46      \\ 
\bottomrule
\end{tabular}
\end{table}

\begin{table}[t]
\centering
\caption{Influence of freeze router matrix or not in the SFT stage for the MoE model.
}
\label{table:freeze}
\begin{tabular}{lcc}
\toprule
 & Frozen & Not-frozen\\
\midrule
 MRPC                     &  91.17&  90.69  
                       \\      
\bottomrule
\end{tabular}
\end{table}

\textbf{The Influence of Freezing Router Matrix.} Based on the observation that LLM's universal world information will impact their performance~\cite{he2023preserving,mukhoti2023finetuning,dou2023loramoe}, we follow \cite{li2023merge} to freeze the router matrix when fine-tuning the MoE model. Here, we compared the empirical results from fine-tuning Switch Transformer on the MRPC dataset in \cref{table:freeze}.

\label{app:levelcompress}



\begin{table*}[t]
\centering
\caption{Efficiency evaluation during the SFT stage.}
\label{tab:efficiency}
\scalebox{0.9}{
\begin{tabular}{lcccccc} 
\toprule
 & Memory (GB) & Runtime (s) & TFLOPs/Model& GFLOPs/MLP Layer & Params (M) & Trainable Params (M) \\
\midrule
Full       & 24.02  & 6697.01$\pm$4.76 & 3.90  & 90.63   &  1,073 & 1,073 \\ 
\midrule
NTK-SAP    & 24.08  & 6687.40$\pm$9.81 & 3.37  & 22.66   & 1,073 & 1,073 \\ 
Structured Pruning  & 15.46  & 5759.46$\pm$9.25 & 3.37 & 22.66 & 620 & 620 \\ 
SVD        & 15.47  & 6562.39$\pm$4.98 & 3.37 & 22.66 &619 &619\\
LoRA       & 13.51  & 6424.75$\pm$4.01 & 2.28 & 31.39 & 1,086 & 123\\
\midrule
\alg~      & 15.46  & 5782.12$\pm$8.90 & 3.37& 22.66 & 620 & 620\\
\bottomrule
\end{tabular}
}
\end{table*}
\begin{table*}[t]
\centering
\caption{Efficiency evaluation during the inference stage. 
}
\label{tab:efficiency_infer}
\scalebox{1}{
\begin{tabular}{lccccc} 
\toprule
 & Memory (GB) & Runtime (s) &TFLOPs/Model& GFLOPs/MLP Layer&  Params (M) \\
\midrule
Full       & 5.38  & 170.33$\pm$0.03 & 1.30 & 30.21 & 1,073 \\ 
\midrule
NTK-SAP    & 5.38  & 168.88$\pm$0.08 & 1.12 & 7.55  & 1,073  \\ 
Structured Pruning    & 4.88  & 163.43$\pm$0.25 & 1.12 & 7.55  & 620 \\ 
SVD        & 3.89  & 176.34$\pm$0.68 & 1.12 & 7.55  & 619 \\
LoRA       & 5.38  & 169.53$\pm$0.05 & 1.30 & 30.21  & 1,073 \\
\midrule
\alg~      & 4.88  & 162.37$\pm$0.14 & 1.12 & 7.55 & 620 \\
\bottomrule
\end{tabular}

}
\vspace{-0.2cm}
\end{table*}

\subsection{Efficiency Evaluation}
\label{sec:efficiency}

In this subsection, we compare the efficiency of our method with LoRA, on the MoE model Switch Transformer (in which MLP takes a higher portion of parameters than dense models). 
We provide the FLOPs and total time during SFT in Table \ref{tab:efficiency}, and those during the inference stage in Table \ref{tab:efficiency_infer}. 
For experiments during the SFT stage, we use the SST2 training dataset for 5 rounds, with a batch size of 32 and a sequence length of 100. 
For experiments during the inference stage, we test the model on SST2 validation set with 10 epochs. We set the config for LoRA on all MLP layers (which is different from our top 8 MoE layers) with rank $r=8$, coefficient $\alpha=32$ (the hyperparameters in LoRA).



Several conclusions could be obtained based on the results:
(\rom{1})~\textbf{NTK-SAP and Hardware Efficiency.} As an unstructured pruning method, NTK-SAP does not translate into hardware efficiency improvements. Despite reducing parameter counts, the nature of unstructured pruning doesn’t align with hardware acceleration capabilities, which are more optimized for structured reductions.
(\rom{2})~\textbf{LoRA's Trade-offs.} While LoRA effectively reduces memory usage during the SFT stage by modifying only a few parameters, it fails to improve efficiency during the inference stage. Even if the overhead from the additional low-rank matrices can be eliminated by integrating them back into the full matrices, the efficiency gains remain limited to the training phase rather than inference.
(\rom{3})~\textbf{Structured Pruning and SVD.} These methods offer a balanced trade-off, with structured pruning and SVD significantly reducing memory and parameter sizes. They also perform well in inference, making them favorable for deployment scenarios where memory and speed are critical.
(\rom{4})~\textbf{\alg's Consistency.} Our approach demonstrates similar efficiency gains as structured pruning and SVD. It reduces memory usage and maintains competitive runtime performance both during training and inference, showing its robustness and suitability for deployment.

\section{Conclusion}
\label{section:conclusion}

In this paper, we propose MLP fusion, a novel one-shot model compression method that utilizes clustering to approximate the NTK of the original PLM.
We demonstrate that the fused MLP can both well approximate the output and attain the closest NTK to the original one compared to other one-shot compression methods. 
At the same time, MLP fusion can be applied to Mixture-of-Experts models and obtain larger gains in space efficiency, underlining its universal effectiveness and scalability in PLMs.
One direct extension of our work is using MLP fusion as an initialization method for distillation. 
Compared to re-training from scratch, we expect the information preserved in the fused MLP can ease the following distillation and speed up the model convergence.
We believe MLP fusion sheds some light on the new paradigm for efficient language model fine-tuning.

\section*{Acknowledgements}

This work is supported by National Science Foundation under Award No. IIS-1947203, IIS-2117902, IIS-2137468, and Agriculture and Food Research Initiative (AFRI) grant no. 2020-67021-32799/project accession no.1024178 from the USDA National Institute of Food and Agriculture. The views and conclusions are those of the authors and should not be interpreted as representing the official policies of the funding agencies or the government.




\bibliographystyle{IEEEtran}
\bibliography{ref}

 





\begin{IEEEbiographynophoto}{Mengting Ai}
is a first-year Ph.D. student at the School of Information Sciences, University of Illinois at Urbana-Champaign. She earned her M.S. in Computer Science from the University of Illinois at Urbana-Champaign in 2023 and her B.S. in Computer Science from Xi'an Jiaotong University in 2022, graduating as an outstanding student. Her research focuses on efficient machine learning, particularly large language model compression. Mengting has published in top-tier conferences such as KDD and actively contributes to the research community, including serving on the AAAI 2024 program committee. Her expertise in NLP and scalable AI systems stems from both academic research and collaborative projects, reflecting a commitment to advancing efficient and scalable AI technologies.
\end{IEEEbiographynophoto}


\begin{IEEEbiographynophoto}{Tianxin Wei}
is a fourth-year Ph.D. student at the University of Illinois at Urbana-Champaign. He earned his B.S. in Computer Science from the School of the Gifted Young at the University of Science and Technology of China in 2020. His research focuses on trustworthy machine learning, with applications in language modeling, information retrieval, and agriculture. He is the recipient of the University Nomination for Apple Scholar in AI/ML 2024, the Amazon Internship Fellowship in 2024, the Conference Presentation Award at UIUC in 2023, NeurIPS Scholar Awards in 2022 and 2023, and the ICML Grant Award in 2023. He has authored more than 20 publications at major conferences (ICML, NeurIPS, ICLR, KDD), and his work has earned the SIGIR Best Paper Honorable Mention in 2021. He has served as a program committee member for leading venues (ICML, NeurIPS, ICLR, KDD, AAAI, WSDM, etc.). His dedication to advancing impactful research was recognized with the Outstanding Reviewer Award for KDD 2025.
\end{IEEEbiographynophoto}


\begin{IEEEbiographynophoto}{Yifan Chen} (Member, IEEE) 
received the B.S. degree from Fudan University, Shanghai, China, in 2018, and the PhD degree in Statistics from the University of Illinois Urbana-Champaign in 2023. He is currently an assistant professor in computer science and math at Hong Kong Baptist University. He is broadly interested in developing efficient machine learning algorithms, encompassing both statistical and deep learning models. He has published several papers in these areas, including ICML, Neurips, KDD, etc. 
\end{IEEEbiographynophoto}


\begin{IEEEbiographynophoto}{Zeming Guo}
is currently pursuing a Master of Science at Cornell University, where his research focuses on advancing language modeling. He earned his Bachelor of Science in Information Science from the University of Illinois at Urbana-Champaign in 2023. He has been working at Amazon Web Services (AWS), where he contributed to the development of large-scale distributed systems. His research interests include large language model reasoning, knowledge distillation, and enhancing the efficiency of deep learning models. He has published first-author papers at the prestigious ICML conference.
\end{IEEEbiographynophoto}


\begin{IEEEbiographynophoto}{Jingrui He} (Senior Member, IEEE) 
is a Professor at School of Information Sciences, University of Illinois at Urbana-Champaign. She received her PhD from Carnegie Mellon University in 2010. Her research focuses on heterogeneous machine learning, active learning, neural bandits, and self-supervised learning, with applications in security, agriculture, social network analysis, healthcare, and finance. Dr. He is the recipient of the 2016 NSF CAREER Award, the 2020 OAT Award, three times recipient of the IBM Faculty Award in 2018, 2015 and 2014 respectively, and was selected as IJCAI 2017 Early Career Spotlight. Dr. He has more than 180 publications at major conferences (e.g., ICML, NeurIPS, ICLR, KDD) and journals (e.g., TMLR, TKDD, JMLR), and is the author of two books. Her papers have received the Distinguished Paper Award at FAccT 2022, as well as Bests of the Conference at ICDM 2016, ICDM 2010, and SDM 2010. Dr. He is a Distinguished Member of ACM, a Senior Member of AAAI and IEEE. She is also the Program Co-chair of IEEE BigData 2023.
\end{IEEEbiographynophoto}

\newpage
\onecolumn

\newpage

\appendices
\crefalias{section}{appendix}

\begin{center}
\setlength{\baselineskip}{1.6\baselineskip}
{\LARGE\bf Supplementary Material for ``\mytitle''}\\
\bigskip
\end{center}

\begin{spacing}{1.5}

\section{Details of experiments}

\subsection{Experimental Setup}

\label{app:experiment_set}

We evaluate the proposed MLP fusion on various downstream NLP tasks and provide a sketch of these tasks in this section.
In addition, we succinctly introduce two intuitive while non-trivial baseline methods, ``Sketching'' and ``MMD'', in \Cref{section:related_work}.
Part of the experiment implementations are borrowed from \cite{chen-etal-2022-empowering,chen-etal-2022-inducer,li2023merge,zhao2024swiftascalablelightweightinfrastructure}.
The code for our algorithms is available at \url{https://github.com/weitianxin/MLP_Fusion}.

All the models in this work are implemented by PyTorch. The experiments are all conducted on one Tesla V100 32 GB GPU. 
For NLU tasks, We fine-tune RoBERTa~\cite{liu2019roberta} with an AdamW \cite{loshchilov2018decoupled} optimizer and use a polynomial learning rate scheduler to make the learning rate linearly decay;
concretely, the learning rate is linearly warmed up from 0 for the first 0.06 epoch. 
The learning rate is searched in the range of \{1e-5, 2e-5,4e-5, 6e-5, 8e-5\}, and the batch size is fixed as $32$.
For NLG tasks, we keep using AdamW optimizer to fine-tune GPT-2 \cite{radford2019language}, and a linear learning rate scheduler with a 500-step warmup duration is used. 
The learning rate is tuned in the same range as above while the batch size is fixed to 8. By default, all the compared methods reduce the MLP intermediate size to 768 or a comparable number of parameters from 3076. The reduction/sketching is performed on the last $8$ layers of the PLM by default. For Clustering, we adopt the K-Means algorithm due to its simplicity and effectiveness. To reduce the random variability in the results, the experiments are all averaged over three runs. 

As for Switch Transformer, the learning rate is searched in the range of \{1e-4,2e-4,3e-4,5e-4,1e-3\}, the batch size within the range of \{16,32,64\}, and the training epoch within the range of \{3,5,10,15,20\}. The details of the AdamW optimizer which is fixed for all datasets are given in table \ref{tab:finetune}.

To ensure comparability across methods, we standardize the parameter count reduction for the experts to around 75\%, which means 25\% of the parameters will be retained. All the methods are performed at the top 8 MoE layers of Switch Transformer. 

\begin{table*}[h!]
\centering
\caption{Fine-tuning hyper-parameters setting for Switch Transformer.}
\label{tab:finetune}
\vspace{0.15in}
\begin{tabular}{lcccc}
\toprule
 & Value  \\
\midrule
Optimizer    & AdamW  \\
Adam $\epsilon$        & 1e-08  \\
Adam $\beta$          & (0.9, 0.98)  \\
warm-up steps & 8\\
weight decay&0.01\\
\bottomrule
\end{tabular}
\end{table*}

\subsection{Runtime of fine-tuning after PLM compression}
\label{app:time}

Since our proposed MLP fusion only differs from the sketching and mmd baselines in initialization, we focus on the runtime evaluation of MLP fusion along with two representative methods, regular fine-tuning and pruning.

For a fair comparison, we intentionally run the two NLU tasks on a cluster server (so that no other processes will compete with the model fine-tuning) with one core of a server CPU (Intel(R) Xeon(R) Gold 6240R CPU @ 2.40GHz) on Ubuntu 18.04.
In this setting, we train the RoBERTa model for 100 steps with batch size 32.

Specifically, on SST2, it will take the model with MLP fusion, pruning, and regular fine-tuning around $6746$, $18066$, $9342$ seconds to finish the training, respectively;
on MNLI, the time cost is around $6956$, $17060$, $18966$ seconds for the training.
We remark the architecture of MLP fusion can accelerate the regular fine-tuning by $30\%$ on SST2, and is even $2.7$ times faster in MNLI, which has longer average sequence length.
As for pruning, although it has a comparable prediction performance in the two tasks, its time cost is no less than regular fine-tuning and is much higher in the more lightweight task SST2, due to some overhead cost from its implementation.

\subsection{The parameter count of SVD}

For SVD, to make the parameters retained for each expert matrix equal, we have:
\begin{align*}
    p_{I}\times k+ k + k\times p &\approx k\times (p_{I}+p)\\
    s\times p_{I}\times p &= spp_{I},
\end{align*}
where $s$ is the parameter rate we retain (25\% here), and $k$ is the number of top-k singular values in SVD. For Switch Transformer, we have $p_{I}=4p$, so $k=\dfrac{4}{5}sp$. For Qwen we have $p_I=\dfrac{11}{16}p$, so $k=\dfrac{11}{27}sp$.

\subsection{Details of the Datasets}

\label{app:dataset_license}
We provide the details of the datasets we used in the experiment along with their license here. The statistics can be found in \cref{tab:data_class,tab:data_gpt}.
\begin{table*}[h!]

    \centering
    \caption{Dataset statistics of fine-tuned classification tasks.}
    \label{tab:data_class}
    \begin{tabular}{lcccc}
    \toprule
       Dataset &Category  & Train size & Test Size & Classes \\
    \midrule   
    SST2 &Sentiment Analysis &67,349&872&2\\
    MRPC & Paraphrase Identification&3,668&408&2\\
    CoLA&Linguistic Acceptability Judgment &8,551&1,043&2\\
    MNLI& Textual Entailment &392,702&9,815&3\\

    \bottomrule
    \end{tabular}

\end{table*}

\begin{table*}[h!]

    \centering
    \caption{Dataset statistics of fine-tuned generation tasks.}
    \label{tab:data_gpt}
    \begin{tabular}{lcccc}
    \toprule
       Dataset &Category  & Train size & Test Size & Average Text Length \\
    \midrule   
    WebNLG& Text Generation &35,426&5,150&24.36\\

    \bottomrule
    \end{tabular}

\end{table*}

\begin{itemize}
    \item SST2\cite{socher2013recursive}: SST2, the Stanford Sentiment Treebank version 2, is a popular dataset for sentiment analysis. It contains movie review sentences labeled as positive or negative, excluding neutral sentences, providing a binary classification task. This dataset is notable for its fine-grained annotation, as it includes sentiment labels for every subphase within the sentence parse trees. SST2 is widely used for training and evaluating models on sentiment analysis, testing their ability to understand nuanced emotional tones in text, with the license of CC0: Public Domain.
    \item MRPC\cite{dolan-brockett-2005-automatically}: 
The Microsoft Research Paraphrase Corpus (MRPC) evaluates models on paraphrase identification by using sentence pairs from online news sources. MRPC is a part of the GLUE benchmark and is valuable for assessing a model's ability to understand and compare semantic content in sentences, especially in semantic analysis tasks. The license of MRPC is unknown.
    \item CoLA\cite{warstadt-etal-2019-neural}: The Corpus of Linguistic Acceptability (CoLA) assesses models' linguistic acceptability judgment. It distinguishes between grammatically acceptable and unacceptable sentences, emphasizing the importance of grammatical understanding in language comprehension and model evaluation. The license for CoLA is not specified.
    \item MNLI\cite{warstadt-etal-2019-neural}: The Multi-Genre Natural Language Inference (MNLI) dataset is a diverse corpus for natural language understanding tasks, focusing on textual entailment. It includes pairs of sentences and challenges models to determine whether the second sentence entails, contradicts, or remains neutral to the first sentence. MNLI's wide range of genres and diverse content makes it a robust benchmark for evaluating models in natural language inference tasks. Most of the data are under the OANC’s license, with the other falling under several permissive licenses, a Creative Commons Share-Alike 3.0 Unported
License, and  Creative Commons Attribution 3.0 Unported Licenses.
\item {WebNLG}\cite{508a2f5efa534d5abf071b219aeef5f3}: This dataset is composed of data/text pairs, where ``data'' is in a format of \emph{(subject, property, object)} triple.
For the train and the validation set, there are nine categories extracted from DBpedia;
while in the test set, there are five extra unseen categories,
which can partially reflect the generalizability of the methods.
The input sequences in the training set contain $1$ to $7$ triples, and the lengths of most sequences are bounded by 50 (as each triple only includes three short phrases). 
The official evaluation script is used in our experiments, and we report BLEU \cite{papineni2002bleu}, METEOR~\cite{banerjee2005meteor} and TER \cite{snover2006study} as the metrics.
\end{itemize}

\section{Derivations omitted in the main text}

\subsection{Computational costs of attention and MLP moduels}
\label{app:attn_mlp_costs}

Condensing FFN sub-layers is critical to obtaining a lightweight pre-trained model.
Besides self-attention sub-layers, FFN sub-layers also take a lot of computation time 
and even become the actual bottleneck when the input sequence length is short.
We will verify this claim through the following derivation.

We first recall the most common setting of an MLP in PLMs.
Taking RoBERTa-base as an example, the hidden dimension is $p=768$ and there are $h=12$ heads in each self-attention module;
the intermediate dimension in MLP is $p_I = 4p = 3072$.
In the self-attention sub-layer, given the length-$n$ input $\mtx X$ we need to first compute the query, key, and value matrix $\mtx Q, \mtx K, \mtx V$,
which takes $3 \cdot n p^2$ operations to perform the linear transform (omitting the bias).
For the core self-attention module, we will at least need $h \cdot 2 n^2 (p/h)$ multiplication operations;
the final linear transform will again take $n p^2$ cost.
The total FLOPs of a self-attention sub-layer are around $4 n p^2 + 2 n^2 p$.

As for the FFN sub-layer, the computational cost is clear: $2 n p p_I = 8 n p^2$.
We can check for regular nlp tasks in which the input length $n$ is bounded by $512$, $8 n p^2$ proves to be larger than $4 n p^2 + 2 n^2 p$, when $p=768$. More specifically, when input length $n<2p$, the computation cost of FFN layers becomes the primary bottleneck. This condition is particularly applicable to modern foundational language models \cite{touvron2023llama}, which often possess a massive hidden size even exceeding ten thousand.

\subsection{Maximum mean discrepancies (MMD)}
\label{app:mmd}

We start with a brief introduction to MMD. 
The expression of MMD between two distributions $P$ and $Q$ is given as
\begin{align}
\text{MMD}(P, Q) 
&= \sup_{\lVert f \rVert_{\m H} \leq 1} \E_{X \sim P}[f(X)] - \E_{Y \sim Q}[f(Y)] \nonumber \\
&= \lVert \E_{X \sim P}[ \varphi(X) ] - \E_{Y \sim Q}[ \varphi(Y) ] \rVert_{\m H},
\end{align}
where $\varphi(\cdot): \m X \to \m H$ is the feature map inducing the kernel function $k(x, y) = \langle \varphi(x), \varphi(y) \rangle_{\m H}$ associated with a reproducing kernel Hilbert space (RKHS) $\m H$.
Through the strong reproducing property of the map $\varphi$, we can rewrite the the squared MMD as
\begin{align}
\text{MMD}^2(P, Q)
  &= \lVert \E_{X \sim P} \varphi(X) - \E_{Y \sim Q} \varphi(Y) \rVert_{\m H}^2 \nonumber \\
  &= \langle \E_{X \sim P} \varphi(X), \E_{X' \sim P} \varphi(X') \rangle_{\m H}
   + \langle \E_{Y \sim Q} \varphi(Y), \E_{Y' \sim Q} \varphi(Y') \rangle_{\m H}
   - 2 \langle \E_{X \sim P} \varphi(X), \E_{Y \sim Q} \varphi(Y) \rangle_{\m H} \nonumber \\
  &= \E_{X, X' \sim P} k(X, X') + \E_{Y, Y' \sim Q} k(Y, Y') - 2 \E_{X \sim P, Y \sim Q} k(X, Y),
\end{align}
which is easier to optimize using back-propagation.

Following the empirical distribution view of MLP, we denote the original MLP as $\mu_w$, a uniform discrete distribution over the rows of the embedding matrix $\mtx W$, and the compressed MLP similarly as $\hat \mu$, an empirical distribution evenly distributed over the rows in matrix $\hat{\mtx W} = [\hat{\mtx W}_1, \hat{\mtx b}_1, \hat{\mtx W}_2] \in R^{c \times (2 p + 1)}$). 
We can then optimize the following problem
\begin{align}
    \min_{\hat{\mtx W}} \text{MMD}^2\left(\mu_{w}, \hat \mu\left( \hat{\mtx W} \right)\right),
\end{align}
from which we can obtain $\hat{\mtx W}$.
As in \Cref{sec:clustering}, we can construct the condensed MLP with $\hat{\mtx W}$ as
\begin{align}
\mtx{\wt H}_M = \frac{p_I}{k} \left[\sigma \left(\mtx X (\mtx W^{(m)}_1)^\T + \mtx 1 (\mtx b^{(m)})^\T\right) \mtx W^{(m)}_2 \right] + \mtx 1 \mtx b_2^\T,
\end{align}
which additionally introduces a factor ${p_I}/{c}$ since expectations rather than sums are involved in MMD.

\subsection{NTK preservation}
\label{app:ntk_preserve}

In the main text, we have made the assumption that $\mtx C \mtx{\wt W} \approx \mtx W$ and $\mtx{\wt H}_C \approx \mtx H$. 
This assumption implies that $\nabla_{\mtx H} f$ can be preserved by $\nabla_{\mtx{\wt H}_C} f_c$, which helps obtain $\dotp{\nabla_{\mtx b_2} f(\mtx X)}{\mathrm{sign}\paren{\nabla_{\mtx b_2} f(\mtx Z)}} \approx \dotp{\nabla_{\mtx{\tilde b}_2} f_c(\mtx X)}{\mathrm{sign}\paren{\nabla_{\mtx{\tilde b}_2} f_c(\mtx Z)}}$.

For the remaining three term, {we first address} $\circled{1} \defeq \dotp{\nabla_{\mtx{\wt W}_2} f_c(\mtx X)}{\mathrm{sign}\paren{\nabla_{\mtx{\wt W}_2} f_c(\mtx Z)}}$: 
\begin{align*}
\circled{1} =& \Tr\brkt{\paren{\nabla_{\mtx{\wt H}_C} f_c(\mtx X)}^\T \mtx{\tilde \sigma}_x  \mtx P \cdot
\mathrm{sign}\paren{ \mtx P \mtx{\tilde \sigma}_z^\T \nabla_{\mtx{\wt H}_C} f_c(\mtx Z)}} \\
=& \Tr\brkt{ \paren{\nabla_{\mtx{\wt H}_C} f_c(\mtx X)}^\T \sigma\paren{\mtx X \mtx{\wt W}_1 + \mtx 1 \mtx{\tilde b}_1^\T} \mtx P \cdot \mathrm{sign}\paren{\mtx P^\T \sigma\paren{\mtx{\wt W}_1^\T \mtx Z^\T + \mtx{\tilde b}_1 \mtx 1^\T} \nabla_{\mtx{\wt H}_C} f_c(\mtx Z)} } \\
\overset{(\rom 1)}{=}& \Tr\brkt{ \paren{\nabla_{\mtx{\wt H}_C} f_c(\mtx X)}^\T \sigma\paren{\mtx X \mtx{\wt W}_1 + \mtx 1 \mtx{\tilde b}_1^\T} \mtx C \mtx C^\T \mathrm{sign}\paren{ \sigma\paren{\mtx{\wt W}_1^\T \mtx Z^\T + \mtx{\tilde b}_1 \mtx 1^\T} \nabla_{\mtx{\wt H}_C} f_c(\mtx Z)} } \\
\overset{(\rom 2)}{=}& \Tr\brkt{ \paren{\nabla_{\mtx{\wt H}_C} f_c(\mtx X)}^\T \sigma\paren{\mtx X \mtx{\wt W}_1 \mtx C + \mtx 1 \mtx{\tilde b}_1^\T \mtx C} \mathrm{sign}\paren{\sigma\paren{\mtx C^\T \mtx{\wt W}_1^\T \mtx Z^\T + \mtx C^\T \mtx{\tilde b}_1 \mtx 1^\T} \nabla_{\mtx{\wt H}_C} f_c(\mtx Z)} } \\
\approx& \Tr\brkt{ \paren{\nabla_{\mtx{H}} f(\mtx X)}^\T \sigma\paren{\mtx X \mtx{W}_1 + \mtx 1 \mtx{b}_1^\T} \mathrm{sign}\paren{\sigma\paren{\mtx{W}_1^\T \mtx Z^\T + \mtx{b}_1 \mtx 1^\T} \nabla_{\mtx{H}} f(\mtx Z)} } \\
=& \dotp{\nabla_{\mtx W_2} f(\mtx X)}{\mathrm{sign}\paren{\nabla_{\mtx W_2} f(\mtx Z)}},
\end{align*}
where equation $(\rom 1)$ above holds since $\mtx P = \mtx C \mtx C^\T$ and the positive diagonal matrix $\mtx P$ will not impact the sign of the matrix elements;
as for equation $(\rom 2)$, the ``copy'' matrix $\mtx C$, as discussed in \Cref{sec:clustering}, is free to be brought inside both the sign function and the activation function.

For $\dotp{\nabla_{\mtx W_1} f(\mtx X)}{\mathrm{sign}\paren{\nabla_{\mtx W_1} f(\mtx Z)}}$, 
we need to verify the product 
$$\mtx X^\T \brkt{\paren{\nabla_{\mtx H} f(\mtx X) \mtx W_2^\T} \odot \mtx \sigma'_x} \cdot \mathrm{sign}\paren{\brkt{\paren{\nabla_{\mtx H} f(\mtx Z) \mtx W_2^\T} \odot \mtx \sigma'_z}^\T \mtx Z}$$ 
can be approximated by
$\circled{2} \defeq \mtx X^\T \brkt{\paren{\nabla_{\mtx{\wt H}_C} f_c(\mtx X) \mtx{\wt W}_2^\T \mtx P} \odot \mtx{\tilde\sigma}'_x} \cdot \mathrm{sign}\paren{\brkt{\paren{\nabla_{\mtx{\wt H}_C} f_c(\mtx Z) \mtx{\wt W}_2^\T \mtx P} \odot \mtx{\tilde\sigma}'_z}^\T \mtx Z}$,
where $\mtx{\tilde\sigma}'_x \defeq \sigma'\paren{\mtx X \mtx{\wt W}_1 + \mtx 1 \mtx{\tilde b}_1^\T}$, $\mtx{\tilde\sigma}_z \defeq \sigma'\paren{\mtx Z \mtx{\wt W}_1 + \mtx 1 \mtx{\tilde b}_1^\T}$, and $\sigma'(\cdot)$ is the derivative of the activation function $\sigma(\cdot)$
\footnote{For simplicity we assume the activation function $\sigma(\cdot)$ is differentiable everywhere.}.
We show the derivation as follows:
\begin{align*}
\circled{2} 
\overset{(\rom 1)}{=}& \mtx X^\T \brkt{\paren{\nabla_{\mtx{\wt H}_C} f_c(\mtx X) \mtx{\wt W}_2^\T} \odot \mtx{\tilde\sigma}'_x} \mtx P \cdot \mathrm{sign}\paren{\mtx P \brkt{\paren{\nabla_{\mtx{\wt H}_C} f_c(\mtx Z) \mtx{\wt W}_2^\T} \odot \mtx{\tilde\sigma}'_z}^\T \mtx Z} \\
=& \mtx X^\T \brkt{\paren{\nabla_{\mtx{\wt H}_C} f_c(\mtx X) \mtx{\wt W}_2^\T} \odot \mtx{\tilde\sigma}'_x} \mtx C \mtx C^\T \cdot \mathrm{sign}\paren{\brkt{\paren{\nabla_{\mtx{\wt H}_C} f_c(\mtx Z) \mtx{\wt W}_2^\T} \odot \mtx{\tilde\sigma}'_z}^\T \mtx Z} \\
=& \mtx X^\T \brkt{\paren{\nabla_{\mtx{\wt H}_C} f_c(\mtx X) \mtx{\wt W}_2^\T \mtx C} \odot \paren{\mtx{\tilde\sigma}'_x \mtx C}} \cdot \mathrm{sign}\paren{\brkt{\paren{\nabla_{\mtx{\wt H}_C} f_c(\mtx Z) \mtx{\wt W}_2^\T \mtx C} \odot \paren{\mtx{\tilde\sigma}'_z \mtx C}}^\T \mtx Z} \\
\approx& \mtx X^\T \brkt{\paren{\nabla_{\mtx H} f(\mtx X) \mtx W_2^\T} \odot \mtx \sigma'_x} \cdot \brkt{\paren{\nabla_{\mtx H} f(\mtx Z) \mtx W_2^\T} \odot \mtx \sigma'_z}^\T \mtx Z,
\end{align*}
in which we obtain equation $(\rom 1)$ because $\mtx P$ as a diagonal matrix has the same scaling effect on the Hadamard product $\brkt{\paren{\nabla_{\mtx{\wt H}_C} f_c(\mtx X) \mtx{\wt W}_2^\T} \odot \mtx{\tilde\sigma}'_x}$ as on one of its component $\nabla_{\mtx{\wt H}_C} f_c(\mtx X) \mtx{\wt W}_2^\T$;
the rest equations simply follow the previous derivations.

For the last term $\dotp{\nabla_{\mtx b_1} f(\mtx X)}{\mathrm{sign}\paren{\nabla_{\mtx b_1} f(\mtx Z)}}$, we solely need to replace the above input matrix $\mtx X, \mtx Z$ with $\mtx 1^\T$, and all the derivation steps will follow.

\subsection{NTK preservation for MoE modules}
\label{app:ntk_preserve_moe}

Considering there are no bias terms in common MoE modules, we solely study the NTK preservation for two weight matrices $\widebar{\mtx W}_1, \widebar{\mtx W}_2$ in this subsection.
In advance of the derivation, 
we first restate the assumptions in \Cref{sec:ntk_verify} that $\wt{\mtx W}_1 \mtx C^{(m)} \approx \widebar{\mtx W}_1$, $(\mtx C^{(m)})^\T  \wt{\mtx W}_2 \approx \widebar{\mtx W}_2$, and $\wt{\mtx H}_m \approx {\mtx H}_m$; 
we also pay close attention to the standalone matrix $\mtx P^{(m)}_i$ for MoE modules:
\begin{align*}
\mathbf{P}^{(m)}_i &= \mathbf{C}^{(m)}\mathbf{R}_i(\mathbf{C}^{(m)})^{\top} \\
&=\begin{bmatrix}
	  \mathbf{C}_{1}[\mtx G(\mathbf{x}_i)]_{1}\mathbf{I}_{p_{I}}\mathbf{C}_{1}^{\top} &    &  \\
	  & ... &   \\
	  &    & \mathbf{C}_{N}[\mtx G(\mathbf{x_i})]_{N}\mathbf{I}_{p_{I}}\mathbf{C}_{N}^{\top}
	 \end{bmatrix}
= \begin{bmatrix}
	  [\mtx G(\mathbf{x}_i)]_{1} \cdot \mathbf{C}_{1} \mathbf{C}_{1}^{\top} &    &  \\
	  & ... &   \\
	  &    & [\mtx G(\mathbf{x_i})]_{N} \cdot \mathbf{C}_{N} \mathbf{C}_{N}^{\top}
	 \end{bmatrix}_{N\cdot c \times N\cdot c}.
\end{align*}
We comment that $\mathbf{P}^{(m)}_i$ similarly is also a positive diagonal matrix;
intuitively, it represents the cluster sizes in each expert, additionally weighted by the gating score in MoE routers.
Moreover, due to the special property of a diagonal matrix, we further have
$$
\mathbf{P}^{(m)}_i = \mathbf{C}^{(m)} (\mathbf{C}^{(m)})^{\top} \cdot \wt{\mtx R}_i,
\quad \text{where~} \wt{\mtx R}_i \defeq \operatorname{diag}(\mtx G(\mtx{x}_i)) \otimes \mtx I_c.
$$
Denoting the compressed MoE model as $\tilde{f}_m$, we start with addressing $\circled{1} \defeq \dotp{\nabla_{\mtx{\wt W}_2} \tilde{f}_m(\mtx X)}{\mathrm{sign}\paren{\nabla_{\mtx{\wt W}_2} \tilde{f}_m(\mtx Z)}}$ as:
\begin{align*}
\circled{1} =& \Tr\brkt{
\paren{\sum_i \paren{\nabla_{\wt{\mtx H}_m} \tilde{f}_m(\mtx X)}_i \mtx{\tilde \sigma}_{x, i}^\T \mtx P^{(m)}_i} \cdot
\mathrm{sign}\paren{\sum_i \mtx P^{(m)}_i \mtx{\tilde \sigma}_{z, i} \paren{\nabla_{\wt{\mtx H}_m} \tilde{f}_m(\mtx Z)}_i^\T }} \\
=& \Tr\brkt{
\paren{\sum_i \paren{\nabla_{\wt{\mtx H}_m} \tilde{f}_m(\mtx X)}_i 
\sigma\paren{\mtx x_i^\T \wt{\mtx W}_1} 
\mtx P^{(m)}_i} \cdot 
\mathrm{sign}\paren{
\sum_i \mtx P^{(m)}_i 
\sigma\paren{\wt{\mtx W}_1^\T \mtx z_i}
\paren{\nabla_{\wt{\mtx H}_m} \tilde{f}_m(\mtx Z)}_i^\T
}} \\
\overset{(\rom 1)}{=}& \Tr\brkt{ 
\paren{\sum_i \paren{\nabla_{\wt{\mtx H}_m} \tilde{f}_m(\mtx X)}_i 
\sigma\paren{\mtx x_i^\T \wt{\mtx W}_1} 
\mtx C^{(m)} \mtx R_i (\mtx C^{(m)})^\T}
\mathrm{sign}\paren{
\mathbf{C}^{(m)} (\mathbf{C}^{(m)})^{\top} \sum_i \wt{\mtx R}_i
\sigma\paren{\wt{\mtx W}_1^\T \mtx z_i}
\paren{\nabla_{\wt{\mtx H}_m} \tilde{f}_m(\mtx Z)}_i^\T}
}.
\end{align*}
The last equation $(\rom 1)$ above holds since $\mtx P^{(m)}_i = \mathbf{C}^{(m)} (\mathbf{C}^{(m)})^{\top} \cdot \wt{\mtx R}_i$.
Considering the positive diagonal matrix $\mathbf{C}^{(m)} (\mathbf{C}^{(m)})^{\top}$ will not impact the sign of the matrix elements, we have
\begin{align*}
\circled{1} =& \Tr\brkt{ 
\paren{\sum_i \paren{\nabla_{\wt{\mtx H}_m} \tilde{f}_m(\mtx X)}_i 
\sigma\paren{\mtx x_i^\T \wt{\mtx W}_1} 
\mtx C^{(m)} \mtx R_i (\mtx C^{(m)})^\T}
\cdot \mathrm{sign}\paren{
\sum_i \wt{\mtx R}_i
\sigma\paren{\wt{\mtx W}_1^\T \mtx z_i}
\paren{\nabla_{\wt{\mtx H}_m} \tilde{f}_m(\mtx Z)}_i^\T}
} \\
=& \Tr\brkt{ 
\paren{\sum_i \paren{\nabla_{\wt{\mtx H}_m} \tilde{f}_m(\mtx X)}_i 
\sigma\paren{\mtx x_i^\T \wt{\mtx W}_1 \mtx C^{(m)}} 
\mtx R_i} \cdot
\mathrm{sign}\paren{
\sum_i (\mtx C^{(m)})^\T \wt{\mtx R}_i
\sigma\paren{\wt{\mtx W}_1^\T \mtx z_i}
\paren{\nabla_{\wt{\mtx H}_m} \tilde{f}_m(\mtx Z)}_i^\T}
} \\
\overset{(\rom 2)}{=}& \Tr\brkt{ 
\paren{\sum_i \paren{\nabla_{\wt{\mtx H}_m} \tilde{f}_m(\mtx X)}_i 
\sigma\paren{\mtx x_i^\T \wt{\mtx W}_1 \mtx C^{(m)}} 
\mtx R_i} \cdot
\mathrm{sign}\paren{
\sum_i {\mtx R}_i 
\sigma\paren{(\mtx C^{(m)})^\T \wt{\mtx W}_1^\T \mtx z_i}
\paren{\nabla_{\wt{\mtx H}_m} \tilde{f}_m(\mtx Z)}_i^\T}
} \\
\approx& \Tr\brkt{ 
\paren{\sum_i \paren{\nabla_{{\mtx H}_m} {f}_m(\mtx X)}_i 
\sigma\paren{\mtx x_i^\T \widebar{\mtx W}_1} 
\mtx R_i} \cdot
\mathrm{sign}\paren{
\sum_i \mtx R_i
\sigma\paren{\widebar{\mtx W}_1^\T \mtx z_i}
\paren{\nabla_{{\mtx H}_m} {f}_m(\mtx Z)}_i^\T}
} \\
=& \dotp{\nabla_{\widebar{\mtx W}_2} f_m(\mtx X)}{\mathrm{sign}\paren{\nabla_{\widebar{\mtx W}_2} f_m(\mtx Z)}},
\end{align*}
where  
equation $(\rom 2)$ holds since 
$$
(\mtx C^{(m)})^\T \wt{\mtx R}_i = {\mtx R}_i (\mtx C^{(m)})^\T,
$$
and the ``copy'' matrix $\mtx C^{(m)}$, as discussed in \Cref{sec:clustering}, is free to be brought inside both the sign and the activation function.

For the rest term $\dotp{\nabla_{\wt{\mtx W}_1} \tilde{f}_m(\mtx X)}{\mathrm{sign}\paren{\nabla_{\wt{\mtx W}_1} \tilde{f}_m(\mtx Z)}}$, we need to show 
$$
\paren{\sum_i
\mtx x_i \brkt{\paren{\mtx R_i \widebar{\mtx W}_{2} (\nabla_{\mtx H_m} f_m(\mtx X))_i} \odot \mtx \sigma'_{x, i}}^{\T}
}  \cdot \mathrm{sign}\paren{\sum_i 
\brkt{\paren{\mtx R_i \widebar{\mtx W}_{2} (\nabla_{\mtx H_m} f_m(\mtx Z))_i} \odot \mtx \sigma'_{z, i}} \mtx z_i^\T}
$$ 
can be approximated by 
$$\circled{2} \defeq \paren{
\sum_i \mtx x_i \brkt{\paren{
\mtx P_i^{(m)} \wt{\mtx W}_2 (\nabla_{\wt{\mtx H}_m} \tilde{f}_m(\mtx X))_i} \odot \mtx{\tilde\sigma}'_{x,i}}^\T 
} \cdot \mathrm{sign}\paren{ \sum_i \brkt{\paren{
\mtx P_i^{(m)} \wt{\mtx W}_2 (\nabla_{\wt{\mtx H}_m} \tilde{f}_m(\mtx Z))_i} \odot \mtx{\tilde\sigma}'_{z,i}} \mtx z_i^\T},
$$
where $\mtx{\tilde\sigma}'_{x, i} \defeq \sigma'\paren{\wt{\mtx W}_1^\T \mtx x_i}$, $\mtx{\tilde\sigma}_{z, i} \defeq \sigma'\paren{\wt{\mtx W}_1^\T \mtx z_i}$, and $\sigma'(\cdot)$ is the derivative of the activation function $\sigma(\cdot)$:
\begin{align*}
\circled{2} 
\overset{(\rom 1)}{=}& \paren{\sum_i \mtx x_i \brkt{\paren{\wt{\mtx W}_2 (\nabla_{\wt{\mtx H}_m} \tilde{f}_m(\mtx X))_i } \odot \tilde{\mtx \sigma}'_{x, i}}^\T \mtx P^{(m)}_i} \cdot 
\mathrm{sign}\paren{\sum_i\mtx P^{(m)}_i\brkt{\paren{\wt{\mtx W}_2 (\nabla_{\wt{\mtx H}_m} \tilde{f}_m(\mtx Z))_i  } \odot \tilde{\mtx \sigma}'_{z, i}} \mtx z_i^\T} \\
=& \paren{\sum_i \mtx x_i \brkt{\paren{\wt{\mtx W}_2 (\nabla_{\wt{\mtx H}_m} \tilde{f}_m(\mtx X))_i } \odot \tilde{\mtx \sigma}'_{x, i}}^\T \mtx C^{(m)} \mtx R_i (\mtx C^{(m)})^\T} \cdot \\
&\qquad \mathrm{sign}\paren{\mtx C^{(m)} (\mtx C^{(m)})^\T \sum_i \wt{\mtx R}_i
\brkt{\paren{\wt{\mtx W}_2 (\nabla_{\wt{\mtx H}_m} \tilde{f}_m(\mtx Z))_i  } \odot \tilde{\mtx \sigma}'_{z, i}} \mtx z_i^\T} \\
=& \paren{\sum_i \mtx x_i \brkt{\paren{\wt{\mtx W}_2 (\nabla_{\wt{\mtx H}_m} \tilde{f}_m(\mtx X))_i } \odot \tilde{\mtx \sigma}'_{x, i}}^\T \mtx C^{(m)} \mtx R_i (\mtx C^{(m)})^\T} \cdot 
\mathrm{sign}\paren{\sum_i \wt{\mtx R}_i
\brkt{\paren{\wt{\mtx W}_2 (\nabla_{\wt{\mtx H}_m} \tilde{f}_m(\mtx Z))_i  } \odot \tilde{\mtx \sigma}'_{z, i}} \mtx z_i^\T} \\
=& \paren{\sum_i \mtx x_i \brkt{\paren{
(\mtx C^{(m)})^\T \wt{\mtx W}_2 (\nabla_{\wt{\mtx H}_m} \tilde{f}_m(\mtx X))_i} \odot \paren{(\mtx C^{(m)})^\T \tilde{\mtx \sigma}'_{x, i}}}^\T \mtx R_i} \cdot \\
&\qquad \mathrm{sign}\paren{\sum_i (\mtx C^{(m)})^\T \wt{\mtx R}_i
\brkt{\paren{\wt{\mtx W}_2 (\nabla_{\wt{\mtx H}_m} \tilde{f}_m(\mtx Z))_i  } \odot \tilde{\mtx \sigma}'_{z, i}} \mtx z_i^\T},
\end{align*}
in which we obtain equation $(\rom 1)$ because $\mtx P^{(m)}$ as a diagonal matrix has the same scaling effect on the Hadamard product $\brkt{\paren{\wt{\mtx W}_2 (\nabla_{\wt{\mtx H}_m} \tilde{f}_m(\mtx X))_i} \odot \tilde{\mtx \sigma}'_{x, i}}$ as on one of its component $\wt{\mtx W}_2 (\nabla_{\wt{\mtx H}_m} \tilde{f}_m(\mtx X))_i$.
Next, we again utilize the relation 
$$(\mtx C^{(m)})^\T \wt{\mtx R}_i = {\mtx R}_i (\mtx C^{(m)})^\T$$ 
and have
\begin{align*}
\circled{2} =& \paren{\sum_i \mtx x_i \brkt{\paren{
(\mtx C^{(m)})^\T \wt{\mtx W}_2 (\nabla_{\wt{\mtx H}_m} \tilde{f}_m(\mtx X))_i} \odot \paren{(\mtx C^{(m)})^\T \sigma'\paren{\wt{\mtx W}_1^\T \mtx x_i}}}^\T \mtx R_i} \cdot \\
&\qquad \mathrm{sign}\paren{\sum_i {\mtx R}_i
\brkt{\paren{(\mtx C^{(m)})^\T \wt{\mtx W}_2 (\nabla_{\wt{\mtx H}_m} \tilde{f}_m(\mtx Z))_i} \odot \paren{(\mtx C^{(m)})^\T \sigma'\paren{\wt{\mtx W}_1^\T \mtx z_i}}
} \mtx z_i^\T} \\
=& \paren{\sum_i \mtx x_i \brkt{\paren{
\mtx R_i (\mtx C^{(m)})^\T \wt{\mtx W}_2 (\nabla_{\wt{\mtx H}_m} \tilde{f}_m(\mtx X))_i} \odot \sigma'\paren{(\mtx C^{(m)})^\T \wt{\mtx W}_1^\T \mtx x_i}
}^\T} \cdot \\
&\qquad \mathrm{sign}\paren{\sum_i 
\brkt{\paren{ {\mtx R}_i  (\mtx C^{(m)})^\T \wt{\mtx W}_2 (\nabla_{\wt{\mtx H}_m} \tilde{f}_m(\mtx Z))_i} \odot 
\sigma'\paren{(\mtx C^{(m)})^\T \wt{\mtx W}_1^\T \mtx z_i}
} \mtx z_i^\T} \\
\approx& \paren{\sum_i
\mtx x_i \brkt{\paren{\mtx R_i \widebar{\mtx W}_{2} (\nabla_{\mtx H_m} f_m(\mtx X))_i} \odot \mtx \sigma'_{x, i}}^{\T}
}  \cdot \mathrm{sign}\paren{\sum_i 
\brkt{\paren{\mtx R_i \widebar{\mtx W}_{2} (\nabla_{\mtx H_m} f_m(\mtx Z))_i} \odot \mtx \sigma'_{z, i}} \mtx z_i^\T},
\end{align*}
and the last equation holds again due to the previous special property of $\mtx R_i$ and $\mtx C^{(m)}$.

\subsection{Model requirements for SGD NTK}
\label{app:sgd_ntk_act}

To preserve the regular SGD NTK, the scale of the weight parameters needs to be adjusted.
We re-define the efficient MLP model as ($f_c, \mtx{\sigma}_x, \mtx{\sigma}_z, \mtx{\tilde\sigma}_x, \mtx{\tilde\sigma}_z$ will also be accordingly re-defined):
\begin{align}
\label{eqn:clustering_mlp_sgd}
\mtx{\wt H}_C \defeq \sigma \paren{\mtx X \mtx{W}^{(c)}_1 + \mtx 1 \paren{\mtx{b}^{(c)}_1}^\T} \mtx{W}^{(c)}_2,
\end{align}
where $\mtx{W}^{(c)}_1 \defeq \mtx{\wt W}_1 \mtx P^\frac12, \mtx{b}^{(c)}_1 \defeq \mtx P^\frac12 \mtx{\tilde b}_1 $ and $\mtx{W}^{(c)}_2 \defeq \mtx P^\frac12 \mtx{\wt W}_1$ incorporate the diagonal scaling matrix $\mtx P$ in \Cref{eqn:clustering_mlp}.

We also require the activation function to have the following property:
\begin{align*}
    \sigma \paren{\mtx A \mtx P} = \sigma \paren{\mtx A} \mtx P,
\end{align*}
for arbitrary non-negative diagonal matrix $\mtx P$, which implies $\sigma(0) = 0$, $\sigma(\cdot)$ is piece-wise linear on $\mb R^+, \mb R^-$, and $\sigma'(x)$ is piece-wise constant ($\sigma(1)$ on $\mb R^+$ and $\sigma(-1)$ on $\mb R^-$);
as an instance, the commonly used Rectified Linear Units (ReLU) function~\cite{fukushima1975cognitron} $\sigma_r(x) = \max\set{0, x}$ can satisfy this requirement.

\Cref{eqn:clustering_mlp_sgd} can keep maintaining the approximation for $\mtx H$ and still $\dotp{\nabla_{\mtx b_2} f(\mtx X)}{\nabla_{\mtx b_2} f(\mtx Z)} \approx \dotp{\nabla_{\mtx{b}^{(c)}_2} f_c(\mtx X)}{\nabla_{\mtx{b}^{(c)}_2} f_c(\mtx Z)}$, 
as we only modify the scale of the weight matrices.
We then follow the derivation in the previous subsection and similar results are obtained

For the remaining three term, again we first address $\circled{1} \defeq \dotp{\nabla_{\mtx{W}^{(c)}_2} f_c(\mtx X)}{\nabla_{\mtx{W}^{(c)}_2} f_c(\mtx Z)}$: 
\begin{align*}
\circled{1} =& \Tr\brkt{\paren{\nabla_{\mtx{\wt H}_C} f_c(\mtx X)}^\T \mtx{\tilde \sigma}_x \cdot
\mtx{\tilde \sigma}_z^\T \nabla_{\mtx{\wt H}_C} f_c(\mtx Z)} \\
=& \Tr\brkt{\paren{\nabla_{\mtx{\wt H}_C} f_c(\mtx X)}^\T \sigma\paren{\mtx X \mtx{W}^{(c)}_1 + \mtx 1 \paren{\mtx{b}^{(c)}_1}^\T} \cdot \sigma\paren{\paren{\mtx{W}^{(c)}_1}^\T \mtx Z^\T + \mtx{b}^{(c)}_1 \mtx 1^\T} \nabla_{\mtx{\wt H}_C} f_c(\mtx Z)} \\
\overset{(\rom 1)}{=}& \Tr\brkt{ \paren{\nabla_{\mtx{\wt H}_C} f_c(\mtx X)}^\T \sigma\paren{\mtx X \mtx{\wt W}_1 + \mtx 1 \mtx{\tilde b}_1^\T} \mtx P^\frac12 \mtx P^\frac12 \sigma\paren{\mtx{\wt W}_1^\T \mtx Z^\T + \mtx{\tilde b}_1 \mtx 1^\T} \nabla_{\mtx{\wt H}_C} f_c(\mtx Z)} \\
=& \Tr\brkt{ \paren{\nabla_{\mtx{\wt H}_C} f_c(\mtx X)}^\T \sigma\paren{\mtx X \mtx{\wt W}_1 + \mtx 1 \mtx{\tilde b}_1^\T} \mtx C \mtx C^\T \sigma\paren{\mtx{\wt W}_1^\T \mtx Z^\T + \mtx{\tilde b}_1 \mtx 1^\T} \nabla_{\mtx{\wt H}_C} f_c(\mtx Z)} \\
\approx& \dotp{\nabla_{\mtx W_2} f(\mtx X)}{\nabla_{\mtx W_2} f(\mtx Z)},
\end{align*}
where equation $(\rom 1)$ holds since $\mtx P^\frac12$, as we require, is free to be brought outside the activation function;
the rest derivation simply follows the counterpart in Appendix~\ref{app:ntk_preserve}.

For $\dotp{\nabla_{\mtx W_1} f(\mtx X)}{\nabla_{\mtx W_1} f(\mtx Z)}$, 
we similarly need to verify the product 
$\mtx X^\T \brkt{\paren{\nabla_{\mtx H} f(\mtx X) \mtx W_2^\T} \odot \mtx \sigma'} \cdot \brkt{\paren{\nabla_{\mtx H} f(\mtx Z) \mtx W_2^\T} \odot \mtx \sigma'}^\T \mtx X$ 
can be approximated by
$\circled{2} \defeq \mtx X^\T \brkt{\paren{\nabla_{\mtx{\wt H}_C} f_c(\mtx X) \paren{\mtx{W}^{(c)}_2}^\T} \odot \mtx{\tilde\sigma}'_x} \cdot \brkt{\paren{\nabla_{\mtx{\wt H}_C} f_c(\mtx Z) \paren{\mtx{W}^{(c)}_2}^\T} \odot \mtx{\tilde\sigma}'_z}^\T \mtx X$.

 $\mtx{\tilde\sigma}'_x \defeq \sigma'\paren{\mtx X \mtx{\wt W}_1 + \mtx 1 \mtx{\tilde b}_1^\T}$, $\mtx{\tilde\sigma}_z \defeq \sigma'\paren{\mtx Z \mtx{\wt W}_1 + \mtx 1 \mtx{\tilde b}_1^\T}$

We then show the derivation as follows:
\begin{align*}
\circled{2} 
=& \mtx X^\T \brkt{\paren{\nabla_{\mtx{\wt H}_C} f_c(\mtx X) \paren{\mtx{W}^{(c)}_2}^\T} \odot \mtx{\tilde\sigma}'_x} \cdot \brkt{\paren{\nabla_{\mtx{\wt H}_C} f_c(\mtx Z) \paren{\mtx{W}^{(c)}_2}^\T} \odot \mtx{\tilde\sigma}'_z}^\T \mtx X \\
\overset{(\rom 1)}{=}& \mtx X^\T \brkt{\paren{\nabla_{\mtx{\wt H}_C} f_c(\mtx X) \mtx{\wt W}_2^\T} \odot \mtx{\tilde\sigma}'_x} \mtx P^\frac12 \mtx P^\frac12 \cdot \brkt{\paren{\nabla_{\mtx{\wt H}_C} f_c(\mtx Z) \mtx{\wt W}_2^\T} \odot \mtx{\tilde\sigma}'_z}^\T \mtx X \\
=& \mtx X^\T \brkt{\paren{\nabla_{\mtx{\wt H}_C} f_c(\mtx X) \mtx{\wt W}_2^\T} \odot \mtx{\tilde\sigma}'_x} \mtx C \mtx C^\T \cdot \brkt{\paren{\nabla_{\mtx{\wt H}_C} f_c(\mtx Z) \mtx{\wt W}_2^\T} \odot \mtx{\tilde\sigma}'_z}^\T \mtx X \\
=& \mtx X^\T \brkt{\paren{\nabla_{\mtx{\wt H}_C} f_c(\mtx X) \mtx{\wt W}_2^\T} \odot \sigma'\paren{\mtx X \mtx{W}^{(c)}_1 + \mtx 1 \paren{\mtx{b}^{(c)}_1}^\T}} \mtx C \mtx C^\T \\
&\qquad \cdot \brkt{\paren{\nabla_{\mtx{\wt H}_C} f_c(\mtx X) \mtx{\wt W}_2^\T} \odot \sigma'\paren{\mtx X \mtx{W}^{(c)}_1 + \mtx 1 \paren{\mtx{b}^{(c)}_1}^\T}}^\T \mtx X \\
\overset{(\rom 2)}{=}& \mtx X^\T \brkt{\paren{\nabla_{\mtx{\wt H}_C} f_c(\mtx X) \mtx{\wt W}_2^\T} \odot \sigma'\paren{\mtx X \mtx{\wt W}_1 + \mtx 1 \mtx{\tilde b}_1^\T}} \mtx C \mtx C^\T \\
&\qquad \cdot \brkt{\paren{\nabla_{\mtx{\wt H}_C} f_c(\mtx X) \mtx{\wt W}_2^\T} \odot \sigma'\paren{\mtx X \mtx{\wt W}_1 + \mtx 1 \mtx{\tilde b}_1^\T}}^\T \mtx X,
\end{align*}
in which we obtain equation $(\rom 1)$ because $\mtx P^\frac12$ as a diagonal matrix has the same scaling effect on the Hadamard product $\brkt{\paren{\nabla_{\mtx{\wt H}_C} f_c(\mtx X) \paren{\mtx{W}^{(c)}_2}^\T} \odot \mtx{\tilde\sigma}'_x}$ as on one of its component $\nabla_{\mtx{\wt H}_C} f_c(\mtx X) \paren{\mtx{W}^{(c)}_2}^\T$;
equation $(\rom 2)$ holds because $\sigma'$ is piece-wise constant and the scaling matrix $\mtx P^\frac12$ will not change the signs of the elements within.
Then, following the same derivation as in the previous derivations,
we have $\circled{2} \approx \mtx X^\T \brkt{\paren{\nabla_{\mtx H} f(\mtx X) \mtx W_2^\T} \odot \mtx \sigma'} \cdot \brkt{\paren{\nabla_{\mtx H} f(\mtx Z) \mtx W_2^\T} \odot \mtx \sigma'}^\T \mtx X$

For the last term $\dotp{\nabla_{\mtx b_1} f(\mtx X)}{\nabla_{\mtx b_1} f(\mtx Z)}$, we can again replace the above input matrix $\mtx X$ with $\mtx 1^\T$, and all the derivation steps will follow.

\section{Bounding the error of MLP output}
\label{app:boundmlpoutput}


 Recall the object of clustering is:
 $$\min\limits_{C} \|\mathbf W - \mathbf C^\T \widetilde{\mathbf W}\|_F^2$$ The assumption can thus be rewritten in a mathematical manner, which is $\|\mathbf W - \mathbf C^\T \widetilde{\mathbf W}\|_F \leq \varepsilon$ with small $\varepsilon$. Assuming $f(\mathbf W, \mathbf{ C^\T \widetilde{W}})=\|\mathbf{W - C^\T \widetilde{W}}\|_F \leq \varepsilon$, we can provide a standard analysis of MLP output (ignoring $b_1$ for simplicity) as follows.

Denoting $\Delta \coloneqq \sigma(\mathbf{X \widetilde{W}_1 C}) - \sigma(\mathbf{X W_1})$ and following the technical assumptions in [4] that $\|\mathbf{{W}_1}\|_2 \leq C_1, \|\mathbf{{W}_2}\|_2 \leq C_2, \|\mathbf{X}\|_F \leq C_X$ and the activation function $\sigma(\cdot)$ is $L$-Lipschitz continuous. 
Further assuming $\sigma(0) = 0$ (the assumptions hold for commonly used activation functions in PLMs, e.g., ReLU and GELU), we first have
\begin{align*}
    \|{\Delta}\|_F \leq L \|{\mathbf{X \widetilde{W}_1 C - X W_1}}\|_F 
\leq L\|\mathbf X\|_F \|{\mathbf{\widetilde{W}_1 C - W_1}}\|_F \leq L C_X \varepsilon.
\end{align*}
We can then bound $\|{ \mathbf{H - \widetilde{H}_C}}\|_F$ as
\begin{align*}
\|{\sigma(\mathbf{X \widetilde{W}_1 C}) \mathbf{C^\T\widetilde{W}_2} - \sigma(\mathbf{X W}_1) \mathbf W_2}\|_F
\leq&\|\sigma(\mathbf{X \widetilde{W}_1 C})(\mathbf{C^\T\widetilde{W}_2 - W_2})\|_F + \|{(\sigma(\mathbf{X \widetilde{W}_1 C}) - \sigma(\mathbf{X W}_1))\mathbf W_2}\|_F    \\
\newline\leq& \|\sigma(\mathbf{X \widetilde{W}_1 C})\|_F\|\mathbf{C^\T\widetilde{W}_2 - W_2}\|_F + \|{\sigma(\mathbf{X \widetilde{W}_1 C}) - \sigma(\mathbf{X W_1})}\|_F\|{\mathbf W_2}\|_2 \\
\newline=& \|{\Delta}+\sigma(\mathbf{X W_1})\|_F\|\mathbf{C^\T\widetilde{W}_2 - W_2}\|_F+\|{\Delta}\|_F\|{\mathbf W_2}\|_2 \\
\newline\leq& (\|{\Delta}\|_F + \|{\sigma(\mathbf{X {W}_1}})\|_F) \cdot \varepsilon + \|{\Delta}\|_F \cdot C_2   \\
\newline\leq& L (C_2 C_X  + C_1 C_X) \cdot \varepsilon + L C_X \cdot \varepsilon^2,
\end{align*}
where we utilize 
\begin{align*}
\|{\mathbf{\widetilde{W}_1 C - W_1}}\|_F, \| \mathbf{C^\T\widetilde{W}_2 - W_2} \| & \leq \| \mathbf W - \mathbf C^\T \widetilde{\mathbf W}\|_F \leq \varepsilon \\
\newline \|{\sigma(\mathbf{X {W}_1}})\|_F = \|{\sigma(\mathbf{X {W}_1}}) - \sigma(\mathbf{0})\|_F 
&\leq L \|\mathbf{X {W}_1}\|_F \leq L \|\mathbf{X}\|_F \|\mathbf{{W}_1}\|_2 = L C_1 C_X,
\end{align*}
for the derivation.
From the bound, we can observe with the well-learned $\mathbf{C}^\T \mathbf{\widetilde{W}}$ from clustering (small $\varepsilon$), the output error ($\|{ \mathbf{H - \widetilde{H}_C}}\|_F$) will also be small.

Moreover, a similar error analysis can also be applied to Adam NTK. The error analysis $\|{ \mathbf{\mathcal K -  \mathcal{\widetilde K}_C}}\|_F$ of NTK kernel $\mathcal{K}$ can be simplified as analyzing $\mathrm{Tr}[\mathbf{A^\T} \mathrm{sign}(\mathbf{B}) - \mathbf{\widetilde{A}^\T} \mathrm{sign}(\mathbf{\widetilde{B}})]$ where $\mathbf A$, $\mathbf B$ represent arbitrary matrices. The precise derivation of the approximation error bound on Adam NTK, considering the additional assumption on $\|\mathrm{sign}(\mathbf B) - \mathrm{sign}(\mathbf{\widetilde{B}})\|_F$ as described in \cite{balles2018dissecting}, is left as future work.



\section{Discussions}
We are not aware of any potential negative societal impacts regarding our work to the best of our knowledge. For all the used data sets, there is no private personally identifiable information or offensive content. 

Regarding future work, beyond the combination with distillation, we also plan to explore practical compression methods in various domains, including speech processing \cite{dong2018speech}, recommender system \cite{geng2022recommendation,wei2022comprehensive}, and graph mining \cite{ying2021transformers,weiaugmentations}. The derivation of a more precise error analysis with regard to the pre-trained model is also a challenging and promising direction.

\section{Performance Comparison of Representative Methods after Task-specific Fine-tuning}
\label{app:methodfinetune}
\begin{table}[H]
\centering
\caption{Accuracy of Representative Methods After Task-specific Fine-tuning on SST2 Validation Set}
\label{table:finetune}
\scalebox{1}{
\begin{tabular}{lc}
\toprule
Method+Task-specific Fine-tuning    & Accuracy       \\ 
\midrule
Sketch           & 91.86          \\ 
Clustering       & 93.35          \\ 
MMD              & 92.43          \\ 
SVD              & 93.01          \\ 
LTH              & 93.42          \\ 
\midrule
MLP Fusion(Ours) & \textbf{93.79} \\ 
\bottomrule
\end{tabular}}
\end{table}
From Table \ref{table:finetune} and Table \ref{table:nlu} in the paper, we can observe that not all methods can benefit from the layer-wise task-specific tuning module. For example, the accuracy of the Sketch method drops from 91.90 to 91.86.
Meanwhile, our proposed MLP Fusion provides a promising starting point for subsequent optimization through NTK approximation. 
By incorporating a layer-wise task-specific tuning module, 
we can further enhance its performance and still achieve the best results compared to all other baseline methods.

\section{Experiment Results on More Benchmark Datasets}
\label{app:morebenchmarks}

\begin{table}[H]
\centering
\caption{Accuracy of Each Baseline Method on STS-B and QNLI Validation Sets with RoBERTa as the PLM}
\label{table:morebenchmarks}
\begin{tabular}{lcc}
\toprule
Method           & STS-B(7k)      & QNLI(105k)     \\ 
\midrule
RoBERTa          & 91.20          & 92.80          \\ 
DistilRoBERTa    & 88.30          & 90.80          \\ 
\midrule
Sketch           & 86.99          & 89.84          \\ 
Clustering       & 88.12          & 90.63          \\ 
LTH              & 87.37          & 90.87          \\ 

\midrule
MLP Fusion(Ours) & \textbf{89.37} & \textbf{91.03} \\ 
\bottomrule
\end{tabular}
\end{table}

Table \ref{table:morebenchmarks} shows the experimental results on two additional data sets QNLI and STS-B within the GLUE benchmark. We can see the proposed method is still able to achieve the best performance over strong baselines.

\section{Performance Comparison between Proposed Method and Maintaining Output/NTK of The MLP Model}
\begin{table}[H]
\centering
\caption{Performance of Proposed Method and Maintaining Output/NTK of The MLP Model}
\label{table:differentmethods}
\begin{tabular}{lc}
\toprule
Methods               & Accuracy \\ 
\midrule
Maintain NTK Gradient & 91.74    \\ 
Maintain Output       & 92.35    \\ 
\midrule
MLP Fusion (Ours)     & 93.23    \\  
\bottomrule
\end{tabular}
\end{table}
\label{app:ntk_maintain}
In Table \ref{table:differentmethods}, we compare two additional baselines that try to maintain the MLP output and NTK of the pre-trained language model. Our NTK approximation method MLP Fusion still achieves the best performance. The loss that attempts to maintain the output with unsupervised data ranked second. The method that tries to maintain the NTK of the MLP model with gradient has the lowest accuracy. This is mainly because the gradient difference in the loss is difficult to minimize since it requires operating second-order derivatives, which can also be time-consuming. Additionally, gathering labeled data for the loss calculation can also be burdensome.


\end{spacing}

\end{document}